\documentclass[lettersize,journal]{IEEEtran}
\usepackage{amsmath,amsfonts}
\usepackage{amssymb}
\usepackage{verbatim}
\usepackage{graphicx}

\usepackage[algoruled,vlined,linesnumbered]{algorithm2e}
\usepackage{xspace}

\usepackage[normalem]{ulem}

\usepackage{xargs}                      %
\usepackage[pdftex,dvipsnames,table,xcdraw]{xcolor}  %
\usepackage[colorinlistoftodos,prependcaption,textsize=tiny]{todonotes}
\usepackage{array}
\newcolumntype{P}[1]{>{\centering\arraybackslash}p{#1}}
\usepackage{pifont}%
\newcommand{\cmark}{\textcolor{OliveGreen}{\ding{51}}}%
\newcommand{\xmark}{\textcolor{red}{\ding{55}}}%
\usepackage{wrapfig}
\usepackage{hyperref}

\usepackage{tcolorbox}
\newtcolorbox{mybox}{colback=gray!5!white,
                     colframe=gray!75!white,
                     fonttitle=\bfseries,
                     boxrule=1pt}

\usepackage{listings}
\usepackage{xcolor}
\usepackage{enumitem}
\usepackage{placeins}

\definecolor{codegreen}{rgb}{0,0.6,0}
\definecolor{codegray}{rgb}{0.5,0.5,0.5}
\definecolor{codepurple}{rgb}{0.58,0,0.82}
\definecolor{backcolour}{rgb}{0.95,0.95,0.92}

\lstdefinestyle{mystyle}{
    backgroundcolor=\color{backcolour},   
    commentstyle=\color{codegreen},
    keywordstyle=\color{magenta},
    numberstyle=\tiny\color{codegray},
    stringstyle=\color{codepurple},
    basicstyle=\ttfamily\footnotesize,
    breakatwhitespace=false,         
    breaklines=true,                 
    captionpos=b,                    
    keepspaces=true,                 
    numbers=left,                    
    numbersep=5pt,                  
    showspaces=false,                
    showstringspaces=false,
    showtabs=false,                  
    tabsize=2
}
\lstdefinestyle{mystyle}{
    backgroundcolor=\color{backcolour}, 
    basicstyle=\ttfamily\footnotesize,
    breakatwhitespace=true,
    breaklines=true,
    showspaces=false,
    moredelim=[is][\textbf]{@}{@},
    escapeinside={(*@}{@*)}, %
}

\usepackage[firstpage]{draftwatermark}

\SetWatermarkText{\parbox{17cm}{\textcopyright 2024 IEEE.  Personal use of this material is permitted.  Permission from IEEE must be obtained for all other uses, in any current or future media, including reprinting/republishing this material for advertising or promotional purposes, creating new collective works, for resale or redistribution to servers or lists, or reuse of any copyrighted component of this work in other works.}}
\SetWatermarkScale{1}

\SetWatermarkColor[gray]{0.5}
\SetWatermarkFontSize{0.3cm}
\SetWatermarkAngle{0}
\SetWatermarkHorCenter{4.1in}
\SetWatermarkVerCenter{0.4in}

\begin{document}
\bibliographystyle{IEEEtran}

\title{%
Task and Motion Planning for Execution in the Real}

\author{Tianyang Pan, Rahul Shome, Lydia E. Kavraki
\thanks{The authors are with the Department of Computer Science, Rice University; ({\tt\footnotesize tianyang.pan@rice.edu, kavraki@rice.edu}).
RS is currently with the School of Computing, the Australian National University ({\tt\footnotesize rahul.shome@anu.edu.au}). The work was supported in part by NSF 1830549, NSF CCF-2336612 and Rice University Funds.
}}

% \markboth{IEEE Transactions on Robotics (T-RO) 2024}%

\maketitle

\newcommand{\tmplan}{\mathbf{T}}
\newcommand{\tmeplan}{\overline{\tmplan}}

\newcommand{\rs}[1]{{\color{red} #1}}

\newcommand{\revision}[1]{{#1}}

\newcommand{\cacceptrevision}[1]{{#1}}

\newcommand{\algterm}[1]{\textnormal{\textbf{#1}}}
\newcommand{\algresult}[1]{\textit{\textbf{#1}}}
\newcommand{\algsc}[1]{\ensuremath{\textnormal{\texttt{{#1}}}}}

\newcommand{\tp}{{\sc TP}\xspace}
\newcommand{\tamp}{{\sc TAMP}\xspace}
\newcommand{\pddl}{{\sc PDDL}\xspace}
\newcommand{\zt}{{\sc Z3}\xspace}
\newcommand{\fd}{{\sc FD}\xspace}
\newcommand{\Scheduler}{Scheduler\xspace}
\newcommand{\scheduler}{scheduler\xspace}
\newcommand{\mrtamp}{{\sc MR-}\tamp}
\newcommand{\finaledits}[1]{{\color{red} #1}}

\newcommand{\policy}{\pi}
\newcommand{\Observations}{\Omega}
\newcommand{\Obsprobs}{O}
\newcommand{\observation}{o}
\newcommand{\mdpdomain}{D_{\mathrm{MDP}}}
\newcommand{\mpdomain}{D_{\mathrm{MP}}}
\newcommand{\tpdomain}{D_{\mathrm{TP}}}

\newcommand{\pick}{\texttt{Pick}}
\newcommand{\place}{\texttt{Place}}
\newcommand{\open}{\texttt{Open}}
\newcommand{\pushpick}{\texttt{Push-Pick}}
\newcommand{\pickpointcloud}{\texttt{Pick-Point-Cloud}}

\newcommandx{\change}[2][1=]{\todo[linecolor=red,backgroundcolor=red!25,bordercolor=red,#1]{#2}}
\newcommandx{\TODO}[2][1=]{\todo[linecolor=blue,backgroundcolor=blue!25,bordercolor=blue,#1]{#2}}
\newcommandx{\improve}[2][1=]{\todo[linecolor=OliveGreen,backgroundcolor=OliveGreen!25,bordercolor=OliveGreen,#1]{#2}}
\newcommandx{\unsure}[2][1=]{\todo[linecolor=Plum,backgroundcolor=Plum!25,bordercolor=Plum,#1]{#2}}
\newcommandx{\thiswillnotshow}[2][1=]{\todo[disable,#1]{#2}}

\newcommand{\objs}{\mathcal{O}}

\newcommand{\Vars}{\mathcal{P}}
\newcommand{\var}{p}
\newcommand{\States}{\mathcal{S}}
\newcommand{\Actions}{\mathcal{A}}
\newcommand{\state}{s}
\newcommand{\action}{a}
\newcommand{\pre}{\texttt{pre}}
\newcommand{\eff}{\texttt{eff}}
\newcommand{\tplan}{T}
\newcommand{\Constraints}{{\Phi}}
\newcommand{\constraint}{\phi}

\newcommand{\Configs}{X}
\newcommand{\config}{x}
\newcommand{\cspace}{\mathcal{X}}
\newcommand{\traj}{\pi}

\newcommand{\WSpace}{\mathcal{Q}}
\newcommand{\wstate}{w}
\newcommand{\WStates}{\mathcal{W}}
\newcommand{\wmap}{{\mathbf{G}}}
\newcommand{\fground}{g}

\newcommand{\trajend}{1}

\newcommand{\Behaviors}{\mathcal{B}}
\newcommand{\behavior}{\mathbf{b}}

\newcommand{\simexec}{\Pi}

\newcommand{\occl}{\epsilon}
\newcommand{\cloud}{\mathbb{C}}
\newcommand{\obj}{o}

\newcommand{\fullname}{Task and Motion Planning for Execution in the Real\xspace}
\newcommand{\acronym}{TAMPER\xspace}

\begin{abstract}

Task and motion planning represents a powerful set of hybrid planning methods that combine reasoning over discrete task domains and continuous motion generation. Traditional reasoning necessitates task domain models and enough information to ground actions to motion planning queries. Gaps in this knowledge often arise from sources like occlusion or imprecise modeling.
This work generates task and motion plans that include actions cannot be fully grounded at planning time. During execution, such an action is handled by a provided human-designed or learned closed-loop behavior. Execution combines offline planned motions and online behaviors till reaching the task goal. Failures of behaviors are fed back as constraints to find new plans.
Forty real-robot trials and motivating demonstrations are performed to evaluate the proposed framework and compare against state-of-the-art. Results show faster execution time, less number of actions, and more success in problems where diverse gaps arise. The experiment data is shared for researchers to simulate these settings. The work shows promise in expanding the applicable class of realistic partially grounded problems that robots can address.
\end{abstract}


\section{Introduction}
\label{sec:intro}

Autonomous robots that operate in realistic environments and accomplish complex task objectives need to be capable of deciding sequences of actions to take and corresponding motions to execute. 
Task and Motion Planning (\tamp) ~\cite{dantam2018incremental, garrett2020pddlstream, toussaint2015lgp} is a popular class of techniques 
that is applied to such problems.
Given a symbolic representation of a task objective, task planning produces an action sequence called a task plan. A grounding process is necessary to generate motion planning queries for each discrete action.
A motion planner can report queries to be infeasible due to geometric constraints such as collisions. A typical \tamp planner~\cite{dantam2018incremental} will feed back such failures as symbolic constraints to compute an alternative task plan, until a valid task and motion plan is discovered.
While being a general and effective paradigm, a prerequisite for this strategy is the existence of knowledge about the symbolic and geometric representations of the world that is sufficient to ground a symbolic action to a motion planning query. 
This information is typically expected from sensing or part of modeling.
This work focuses on task and motion planning problems where not enough information is available to complete every grounding step at planning time.

\begin{figure}[t!]
    \centering

    \includegraphics[width=0.9\linewidth]{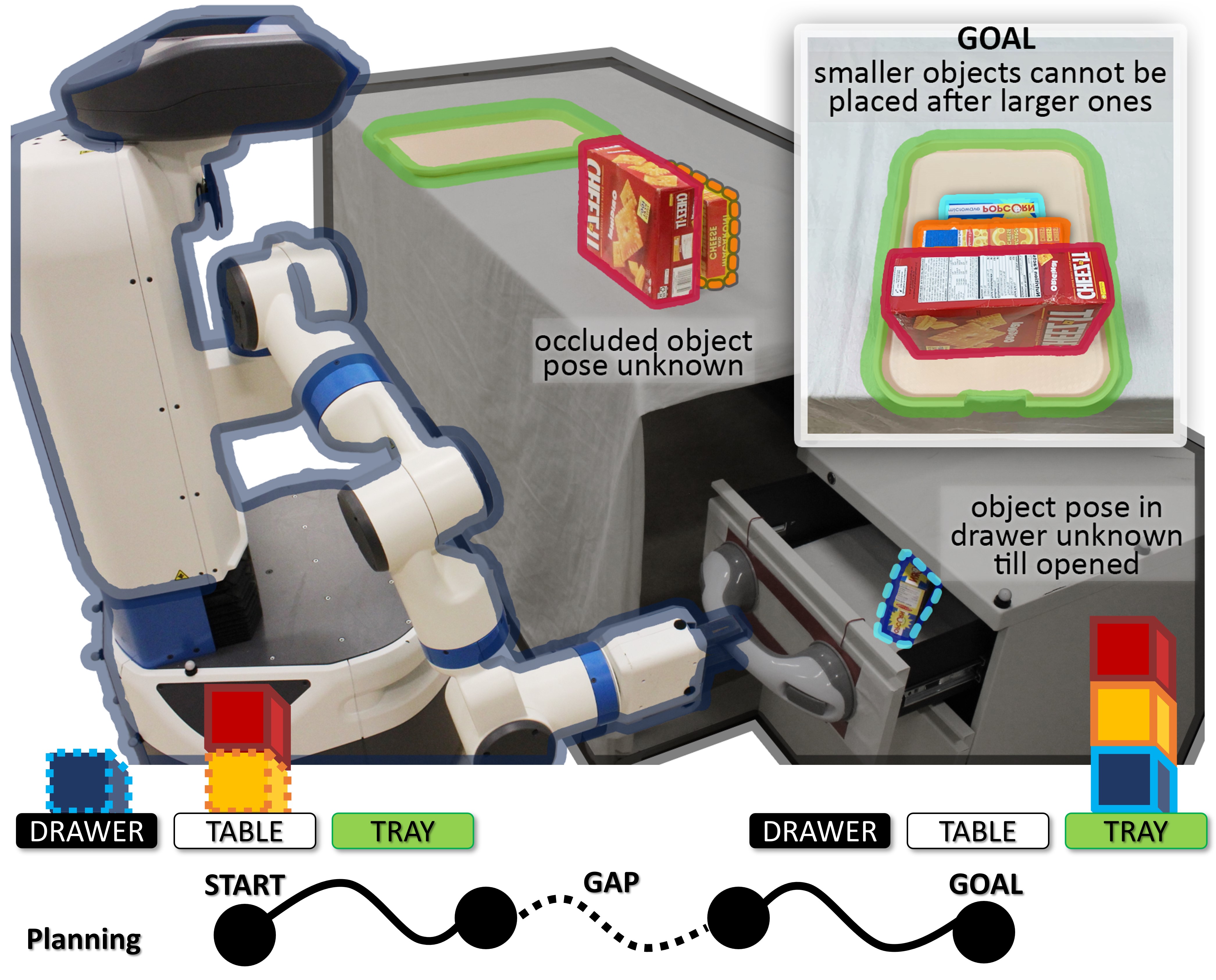}
    
    \caption{The task is to move three objects to the goal tray. The full states of the dotted objects are not known precisely. A task and motion plan will have gaps which can only be filled in during execution, e.g., the blue object's pose can only be known after opening the drawer. The problem imposes constraints visualized in a block-world-like diagram at the lower part of the figure.
    }
    \vspace{-0.2in}
    \label{fig:gaps}
\end{figure}

\begin{figure}[t!]
    \centering
    
    \includegraphics[trim={0cm 5.5cm 0cm 0cm},clip,width=0.94\linewidth]{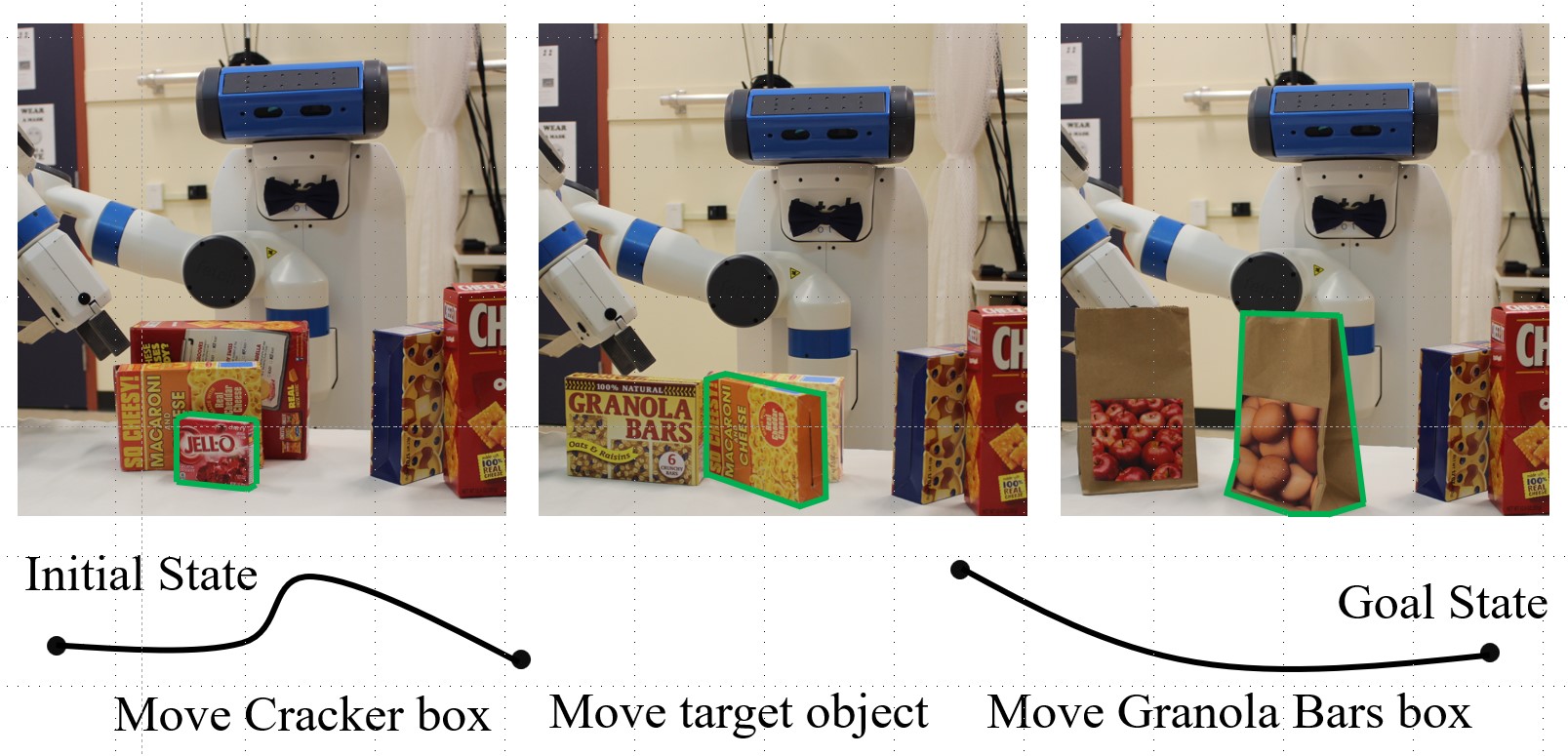}

    \vspace{-0.15in}
    \setlength{\tabcolsep}{0pt}
    \noindent\begin{tabular}{P{0.99in}P{1.11in}P{0.97in}}
    {\scriptsize Object not seen}&{\scriptsize Push simulation incorrect}&{\scriptsize Bag contents unknown} 
    \end{tabular}
    
    \vspace{-0.05in}
    \caption{The source of gaps can be diverse, including ({from left to right}) occlusions, imprecise simulations, and object modeling gaps (two paper bags look the same but contain different groceries).}

    \vspace{-0.2in}
    \label{fig:gapssource}
\end{figure}

A variety of realistic situations introduce partial knowledge of the world for a given task, violating the assumptions of the state-of-the-art \tamp solvers.
For example, as in Fig.~\ref{fig:gaps}, we may know that two objects are on the table and one object is in the drawer, but we do not know the geometric poses of some of the objects due to occlusions, inaccurate sensors, etc. The object pose is necessary to generate robot grasping configurations and motions that achieve there. Traditional \tamp approaches struggle to address problems that might need to ground a \pick\text{ }action that needs to interact with an unknown object. 
Therefore, a discrete plan from the task planner may consist of actions that can be fully grounded to continuous motions through motion planning, as well as actions that \textit{cannot} be grounded due to lack of enough knowledge of the world. Partial grounding creates \textit{gaps} in the resulting task and motion plan. The gaps may arise due to various reasons including perception uncertainties, imprecise simulations, modeling gaps, etc., as illustrated in Fig.~\ref{fig:gapssource}. 
We observe that actions that could not be grounded at planning time might be necessary for the task and could be reasoned during execution time.
This work focuses on how to address scenarios where only partial grounding is possible, and proposes to extend \tamp with closed-loop behaviors that are deployed during execution.

Recent advances~\cite{garrett2020online} focus on \tamp with sensing uncertainties, and have proposed a framework that integrates a \tamp solver more closely into the typical Sense-Plan-Act loop, leveraging sensing and execution to gain more information.
The motion planning difficulty is deferred till execution.
Recent work~\cite{curtis2022M0M} tackles \tamp problems where underlying models of the problem are incomplete, by repeatedly executing one or more steps of the current task and motion plan, gathering new information, re-formulating a new \tamp problem, and replanning.
Other types of gaps such as stochastic execution outcomes are not considered.
The sources of gaps are diverse and might only be known for parts of the \tamp domain.
\cacceptrevision{We focus on such real-world problems in which we are aware that some types of knowledge gaps may exist.}
\textit{This propels our study into maintaining the benefits of coupled task and motion reasoning, while addressing incomplete \tamp domains with gaps that can be filled in during execution.}

\begin{figure}[t!]
    \centering
    
    \includegraphics[width=0.99\linewidth]{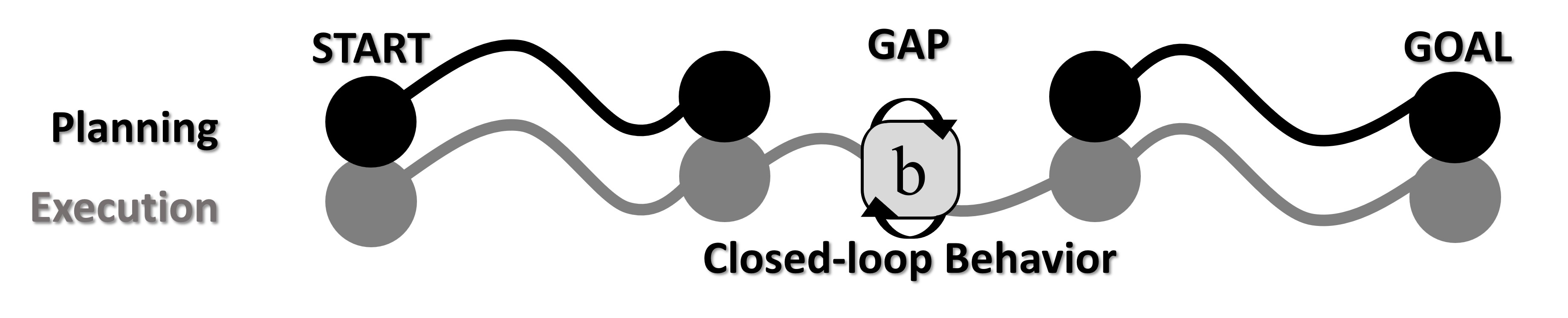}
    \vspace{-0.1in}
    \caption{
        This work introduces behaviors to bridge gaps in the task and motion plan during execution time, and recovers constraints from the task, motion, and execution domains.  
        }

    \vspace{-0.2in}
    \label{fig:gap-behavior}
\end{figure}

We note that \tamp solvers are very powerful for problems where enough information is at hand. In motivating scenarios like Fig.~\ref{fig:gaps}, most of the problem might be known and possible to ground (e.g., the position of the table and the robot).
We propose to build a framework that can leverage existing \tamp solvers for the part of the task where we have good enough information, and leave gaps in the plan for actions that cannot be fully grounded during planning (Fig.~\ref{fig:gaps}).
The output of the planning phase is a partially grounded task and motion plan with gaps. 
This is sent to the execution layer, where we leverage provided closed-loop \textit{behaviors}, as shown in Fig.~\ref{fig:gap-behavior}. These behaviors can be human-designed modules, learning-based local policies, etc., that can be triggered during execution to ground individual actions where gaps exist.
Even if a behavior is carefully designed, its execution may still fail. Our framework allows feedback from such failures as execution-layer constraints.
Overall, the framework proceeds in an incremental fashion computing task plans till a partially grounded sequence of feasible motions plans and behaviors is generated. Behavior failures during execution trigger generation of alternative task plans, till the robot successfully accomplishes the task in execution using some combination of precomputed motion plans and behaviors.   

The primary contribution of the current work is to design a general and extensible framework for \textbf{\fullname \text{} (\acronym)} that can address (a) dealing with symbolic states and actions which might only be partially grounded (creating \textit{gaps}), (b) bridging the \textit{gaps} using closed-loop behaviors comprising human-designed steps or learned controllers, and (c) recovering task-level constraints from the execution feedback from behaviors.
We evaluate our proposed framework extensively on \textit{real-world benchmarks} to demonstrate the efficacy of combining model-based \tamp reasoning with closed-loop behaviors. Our method succeeds more often, and takes fewer actions in real world settings with partial groundings, compared against a baseline implemented based on~\cite{curtis2022M0M}.
We also share a dataset showcasing real-robot \tamp problems that include problem definitions, recorded online data, and instructions to replicate experimental settings.

\section{Related Work}
\label{sec:related-work}
\subsection{Task and Motion Planning}

There are a variety of approaches proposed for integrated \tamp~\cite{garrett2021review}.
A typical approach~\cite{dantam2018incremental} is to handle task planning using a general-purpose satisfiability modulo theories (SMT) solver~\cite{moura2008z3} that supports incremental solving to efficiently consider motion constraints as feedback from motion grounding failures.
Some works leverage black-box samplers encoded as streams to efficiently compute feasible plans~\cite{garrett2020pddlstream}.
\tamp has also been formulated as an optimization problem in~\cite{toussaint2015lgp}. Multi-modal motion planning (MMMP) methods represent the underlying discrete structure of \tamp problems as mode families, and compute the mode transitions as well as the single-mode motions~\cite{hauser2011mmmp, kingston2020informing, kingston2022tro}.
The above methods usually assume perfect sensing and deterministic execution outcomes, and focus on planning with complete knowledge of the world.
If these assumptions do not hold,
such strategies exhaust all possible discrete plans without finding a feasible one.

\tamp in stochastic domains is gaining more and more interest recently. Some works formulate the stochastic \tamp problem as a Markov Decision Process (MDP)~\cite{shah2020anytime, wells2021finite}, assuming a known transition model.
In~\cite{pan2022failure}, the problem of \tamp with failing executions is modeled as an MDP with unknown transition probabilities, where a Beta-Binomial model is leveraged to maintain the belief of the unknown probabilities to minimize the expected number of actions executed to finish tasks.
These works focus on stochastic execution outcomes and assume perfect sensing.

There are approaches that address underlying uncertainty by formulating partially-observable TAMP domains and planning in a belief space.
The general formalism of Partially Observable Markov Decision Processes (POMDPs)~\cite{kaelbling1998pomdp} provides a way to compute optimal policies, but is typically computationally expensive. Efficient online, approximate variants~\cite{somani2013despot, kurniawati2016online} have been proposed. Despite this, careful modeling of state and action spaces is necessary to both reasonably model the problem and maintain reasonable performance.
The fully defined robotic task and motion domain presents tractability issues since the dimensionality of the belief space grows with the cardinality (or size) of the state space.

There have been works that leverage or extend these online POMDP solvers and apply them to robotics manipulation domains~\cite{li2016act, xiao2019icra}. They focus on reasoning over discrete actions choices, while computing continuous motions for high-DoF manipulators is not addressed, in contrast to TAMP literature whose forte is the coupled discrete-continuous problem.
In order to tackle TAMP problems with uncertainty,
different representations and approximations have been proposed for the belief over the partially observable object poses or robot joint values, including Kalman filtering, particle filtering, or Nonparametric Belief Propagation~\cite{garrett2020online, kaelbling2013belief, adu-bredu2021shy-cobra}.
These are mostly closed-loop TAMP methods with the focus on policy computation that needs to react on the fly to new observations with bounded horizon reasoning.
Partially observable \tamp has also been integrated with trajectory-optimization-based techniques~\cite{phiquepal2019opt-po-tamp}.
Strong assumptions are often necessary to simplify problem modeling~\cite{garrett2020online, kaelbling2013belief, adu-bredu2021shy-cobra, phiquepal2019opt-po-tamp}, which allows the resulting POMDP to be solved more efficiently.
For example, the probabilistic model of transitions can be determinized to compute plans in a tractable way~\cite{garrett2020online, kaelbling2013belief}.
Moreover, all the above works typically focus on a single type of gap at a time in terms of the observation model, for instance occlusion or noise in object pose estimation~\cite{garrett2020online, kaelbling2013belief, adu-bredu2021shy-cobra, li2016act, xiao2019icra}, or partially observable attributes of objects~\cite{phiquepal2019opt-po-tamp} (e.g., the color of the opposite side of a cube cannot be seen).

It is important to point out that many types of partial groundings (gaps) could exist in \tamp problems. This again raises open problems in modeling and scalability.
Moreover, it remains to be explored, how long-horizon task-level logical constraints can be enforced in manipulation tasks using a POMDP model, and it is unclear how to represent and use geometric constraints coming from continuous motion computation and execution within in a POMDP framework.
Notably, TAMP on its own cannot address many of these complexities, and the proposed work pushes the traditional
constraint-based TAMP strategy that handles long-horizon, discrete-continuous, high-dimensional search spaces into more realistic partially-grounded scenarios.
We address such a new class of problems by using provided behaviors that deal with various types of local gaps to efficiently reason about planning and execution.

A recent related work~\cite{curtis2022M0M} plans with an incomplete model of the world (e.g., unmodeled objects and occlusion). It proposes to estimate and reason over the properties and affordances of the objects. It keeps replanning with the detected objects in the now until it reaches the goal or a dead end. Our work can deal with not only gaps coming from perception, but also other general types of gaps (e.g., stochastic execution outcomes), as long as specialized execution-level behaviors are provided.
Our main contribution is a general formulation and framework for \tamp with partial groundings, while the necessary engineering effort for each particular problem is left modularized and flexible as input behaviors.

\subsection{Reactive Manipulation}

Substantial research effort has been devoted to developing methods that are reactive and adaptive to external interventions in real-world robotic tasks.
Reactive synthesis approaches~\cite{he2019efficient-symbolic-reactive-synthesis-for-finite-horizon-tasks, wells2021finite} aim at fully computing a policy over the abstraction represented by some formal language such as Linear Temporal Logic (LTL)~\cite{de2013LTLf} to handle human intervention or environmental stochasticity during execution time.
Although these methods have been used for high-DoF robot manipulation tasks,
searching for the abstract policy is already highly complex; so they usually simplify the computation of low-level motions.
Synthesizing such a policy requires full knowledge of the task and motion domain, and when the domain changes the policy must be recomputed. To more efficiently react to environmental model changes, another approach partially constructs the search graph of the task domain and exploits the modularity of Behavior Trees (BTs)~\cite{colledanchise2018BT} as the execution layer to manage the given low-level control policies~\cite{li2021reactive}.
In~\cite{migimatsu2020dynamic}, \tamp is formulated as an optimization problem over relative object poses, and reactive controllers can be leveraged to adapt to real-world changes. It is assumed that the controllers can always react to the disturbances during execution. Hence recomputing a new task plan is not necessary.
All the above synthesis methods mostly compute a policy for the full task given, which becomes computationally prohibitive when considering execution-level constraints or solving long-horizon tasks. We focus on problems where, despite the problem not being fully grounded during planning, we can efficiently reason over motion constraints and execution failures.

Plan execution monitoring methods~\cite{giacomo1998monitoring} specifically focus on detecting the difference between world states and expected states, and recovering from the discrepancies. There have been works that consider execution failures by diagnosing broken parts of the robotic system with either fully observable~\cite{HANHEIDE2017AI} or partially observable state spaces~\cite{corhlu2022explainable}.
Another work reacts to temporal or resource constraints that emerge at execution time~\cite{lemai2004temporal}.
These methods usually do not focus on the combined task and motion planning problem for high-DoF robots, and mostly only consider cases where the \textit{gap} of grounding comes from imperfect simulation model (malfunctioning robot or unmodeled physics) of the real world.
In this work, we focus on the general problem of partial groundings in \tamp, where diverse types of \textit{gaps} can arise.

Another line of research leverages suitable execution-level architectures to achieve reactive and robust behaviors. BTs have become popular as a control structure for complex robot systems because they allow modularity, hierarchy, and feedback control~\cite{colledanchise2018BT, li2021reactive, ogren2022behavior}. Back chaining from the goal condition for planning has been used to create and update BTs to react to external disturbances~\cite{colledanchise2019reactive}. Another work proposes Robust Logical-Dynamical Systems (RLDS) as an alternate structure that has equivalent expressiveness as BTs and algorithms to compose and execute an RLDS~\cite{Paxton2019RLDS}.
These works usually defer motion planning to execution time. In constrast, we leverage the power of \tamp planners that can discover some motion or geometric constraints (e.g., infeasible motions from certain states) at the initial planning stage, before even entering the expensive real-world execution stage.
As pointed out in Sec.~\ref{sec:intro}, we consider cases where even the input behaviors are not guaranteed to complete an action, making feedback of constraints from execution layer and replanning inevitable and necessary --- something that is typically not addressed in many of the above methods.

In another work~\cite{Pezzato2023tro}, active inference approaches are integrated with BTs to enhance the robustness of the given nominal plan (in the form of BT) to unexpected disturbances and partial observability. The primary focus is to enhance the given offline nominal plan during runtime, while our approach can efficiently leverage such behaviors with the proposed task and motion planning and execution framework.

Many of the works in this section are complementary to our proposed framework.
\acronym can efficiently reason over long-horizon task and motion planning problems where partial groundings exist due to lack of full knowledge of the real-world domain. We leverage provided behaviors (see definition in Sec~\ref{sec:formulation:tamper}), which can be in the form of these powerful online local recovery policies, to ground individual actions where grounding gaps exist.

\subsection{Learning-based Methods}
Machine learning techniques have been leveraged to learn abstract hierarchical policies for various robotic tasks from human-designed reward functions or expert demonstrations~\cite{xu2018ntp}, in the form of Finite State Machines or BTs~\cite{french2019lfd}. However, they usually consider motion planning as an online module that does not fail. Since the completeness of planning for high-DoF robotic arms in complex scenes cannot be guaranteed, we inherently reason over motion planning failures as constraints in line with other typical \tamp methods.

Some learning-based methods aim at learning the grounding itself, applied to recover symbols
that permit planning using skills~\cite{gopalan2020simultaneously} or LTL expressions~\cite{patel2020grounding}.
Our focus is on problems where grounding functions might exist but cannot be applied due to uncertainty or insufficient information.

Some works combine Reinforcement Learning with \tamp planners to adapt to unseen environmental dynamics with mobile robots~\cite{jiang2019iros}, or use unsupervised learning to collect data and predict feasibility of task plans in domains where execution may fail~\cite{NoseworthyBrandMoses-RSS-21}. To address problems where the mapping between symbolic abstraction and real-world sensor data is hard to define, the approach in~\cite{kase2020transferable} learns both the high-level model to predict the symbolic state, and the low-level control policies for execution.
Some work looks into the problem of aquiring neuro-symbolic skills given symbolic predicates through learning from demonstration~\cite{silver2022learning}.
To help with gaps in grounding unseen objects, another work aims at learning object-centric representations from RGB images for specific manipulation tasks, where the execution is composed with given primitive skills~\cite{yuan2021sornet}.
There also exist several deep Reinforcement Learning approaches that let the robot learn skills for manipulation~\cite{ibarz2021deepRL} like screwing the cap of a bottle~\cite{levine2016end}, etc. Acquiring such skills is not the focus of this paper. Our framework can readily 
Learning-based methods are hard to scale to general longer-horizon manipulation tasks that \tamp planners excel at solving~\cite{ibarz2021deepRL}. Also, simulation data is necessary because acquiring a large amount of real-world data for robotic manipulation tasks is difficult and expensive. But a simulator might not be a perfect model of the real world,  presenting challenges in applying the policies learned in simulation to real robots, known as the sim-to-real gap~\cite{zhao2020sim2real}. Some methods address this problem, but mostly only focus on recovery from failures caused by unmodeled dynamics, physical interactions, or system errors~\cite{pan2022failure, corhlu2022explainable}.

Again, many of the powerful learning-based local policies in this subsection can be leveraged as input behaviors for \acronym to ground individual actions where grounding gaps exist.

\section{Problem Formulation}
\label{sec:formulation}
We introduce the typical formulation of TAMP in sec.~\ref{sec:formulation:tamp}, and then extend the formulation to the problem we aim to solve in sec.~\ref{sec:formulation:tamper}.

\subsection{Task and Motion Planning}
\label{sec:formulation:tamp}

\noindent\textbf{Task Planning: }Task planning is defined over a domain $\tpdomain$ that consists of

\begin{itemize}
    \item a finite set of states $\States$, where each state $\state \in \States$ is defined over a set of boolean state variables $\Vars = \{\var_0, \var_1, ..., \var_n\}$, i.e., $\state \in 2^{\Vars}$.
    \item a finite set of actions $\Actions$ that allow transition from one state that satisfies the action's pre-conditions (denoted as $\pre(\action) \subseteq \States$) to another state that satisfies the action's end-effect (denoted as $\eff(\action) \subseteq \States$).
    \item a finite set of constraints $\Constraints$, where each constraint $\constraint$ is a logical assertion indicating an action cannot be taken from a set of states, i.e., $\{\state_1, \state_2, ...\}\Rightarrow \neg \action$. Please refer to \cite{dantam2018incremental} for details.
    \item an initial state $\state_{init} \in \States$,
    \item and a finite set of goal states $\States_{goal} \subseteq \States$.
\end{itemize}

The solution to a task planning problem is a task plan $\tplan$ which is a sequence of actions $\action_0, \action_1, ..., \action_l$. Each action $\action_i$ transitions from its state $\state_i \in \pre(\action_i)$ to $\state_{i+1} \in \eff(\action_i)$, where $\state_0 = \state_{init}$ and $\state_{l+1} \in \States_{goal}$.
The task plan must also satisfy all the constraints. Formally, $\forall i, \forall \constraint \in \Constraints, \action_i \vDash \constraint$.

\noindent\textbf{Motion Planning:}
The motion domain $\mpdomain$ describes a $d$ degree of freedom (DoF) robot defining a configuration space $\cspace$, a subset of which is valid $\cspace_{\text{free}} \subseteq \cspace \subset\mathbb{R}^d$.
A trajectory is a time parameterized curve $\traj : [0,\trajend] \rightarrow \cspace_{\text{free}}$, which is called a solution to a motion planning problem between a start $\config_0 \in\cspace$ and a goal $\config_1 \in\cspace$ if $\traj(0) = \config_0, \traj(\trajend) = \config_1$.
It is typically assumed that it suffices to find a geometric path $\traj_i$ for the robot to ground an action $\action_i$ into robotic motions. The validity of these motions, the corresponding $\cspace_{free}$ depend on the geometric components of the workspace like static obstacles and poses of movable objects.

\noindent\textbf{Task and Motion Planning (TAMP):} 
An instance of a TAMP problem includes an initial state $\state_{0}$ and a set of goal states $\States_{goal}$.
TAMP is the problem of finding a feasible $n$-step task and motion plan $\tmplan = (\langle a_{0}, \pi_{0}\rangle, ..., \langle a_{n}, \pi_{n}\rangle)$, where $a_{0}, ..., a_{n}$ is the task plan that satisfies the task planning domain and successfully transitions 
to a goal state $\state_{n} \in \States_{goal}$.
Each discrete action $a_{i}$ transitions from state $\state_{i}$ to the result state $\state_{i+1}$ and corresponds to a feasible motion plan $\traj_{i}$. 

Notably, the geometric feasibility depends on not only the static geometric parts of the environment but also task-dependent aspects like object or environment interactions or task-specific properties or constraints.

\noindent\textbf{Execution after Planning:} 
Once a feasible task and motion plan is computed, in open-loop execution the motion plan can then be sent to controllers which move the robot in the real world. This executes the computed plan. 

Such open-loop execution succeeds on the real robot when the computation and planning used models of the world that closely matches the real world and the robot controller is accurate. A successful open-loop execution is expected to show little deviating from the computation of the planner, as illustrated in Fig.~\ref{fig:all-gaps}a.

The focus is shifted to scenarios where the computation and open-loop execution of task and motion plans can prove insufficient to successfully complete the task during robot execution. In contrast to open-loop execution, interleaving planning and execution alongside combined reasoning across both presents a more powerful paradigm.
\emph{We formulate the problem where both planning and execution for task and motion goals are under consideration.}

We are interested in TAMP problems where we do not have enough information of the world, or in other words, the assumption of a perfect model of the world does not hold.
In such cases, some of the symbolic states and actions may remain partially grounded, and real-world execution is necessary to obtain more information to fully ground and execute such actions.

\subsection{TAMP for Execution}
\label{sec:formulation:tamper}
In this section we define what partial grounding means in our context, and proceed to expand on the formulation of \tamp to involve executions.

\noindent\textbf{World State Space:}
A world state is a vector of all the attributes that models the world inside planning. This goes beyond the degrees of freedom of the robot configuration and can include other properties of the world relevant to the problem.   
The world state space $\WStates$ can include different continuous or discrete state spaces $\WStates = \cspace_\mathrm{robot} \times \cspace_\mathrm{object_1} \times ... \times \cspace_\mathrm{property_1} \times ...$, where $\cspace_\mathrm{robot}$ is the configuration space of the robot, $\cspace_\mathrm{object_1}$ typically describes the SE(3) pose of an object, and $\cspace_\mathrm{property_1}$ can describe a property of the world that is relevant to the task (e.g., whether the table is clean or dirty, or whether the drawer is open or closed).

\noindent\textbf{Grounding Symbolic States and Actions:}
Grounding is the value assignment of all the symbolic variables in an expression.
In the context of TAMP, each symbolic state $\state$ can be fully grounded to a set of attributes describing the world state. This grounding operation can create a one-to-many mapping to a set of (usually infinitely many) possible world states, 
$\wmap(\state_i)\subset \WStates$. Fully grounding a symbolic state $\state_i$ means the state is mapped to a single world state $\wstate_i$.

In a task plan, from a state $\state_i$ that is grounded to $\wstate_i$, grounding a symbolic action $\action_i$ describes how the robot reaches the state $\state_{i+1}$ that follows.

To compute a trajectory for the transition $\langle \state_i, \action_i, \state_{i+1} \rangle$ using a typical motion planner, the grounding process first constructs a motion planning query.
The start configuration of the query has been determined by the configuration $\config$ that is part of $\wstate_i$.
Specifically, $\config$ is the configuration where the robot starts the task (i.e., $i = 0$), or is the end of the preceding trajectory,  $\traj_{i-1}(1), i > 0$. 
The goal configuration can be computed from the definition of action $\action_i$, the configuration space constraints (e.g., opening a drawer can require a specific range of orientation of the gripper), and the set of world states $\wmap(\state_{i+1})$ that the system must to transition into, constrained by the symbolic transition itself.
When such a query can be constructed, a sampling-based motion planner can generate $\traj_i$.
As the robot moves along $\traj_i$, the corresponding world state also evolves.
In the presence of inaccurate or incomplete modeling of the world used during planning, the motion planning query might not be possible to construct. When computed and executed in the real world, such a motion plan $\traj_i$  can be unsafe or may achieve an end world state that is different from the planned $\wstate_{i+1}$. If precomputed plans continue execution undeterred, differences between what was planned and what is executed will propagate along $\traj_j, j>i$. 

For instance, the typical process of grounding a \pick\text{} action involves computing a grasp pose of the target object, then computing the goal configuration that achieves the grasp pose, and finally performing motion planning to this goal. When this process is part of a task and motion plan, the end world state after the trajectory $\traj_i$ should satisfy $\wstate_{i+1} \in \wmap(\state_{i+1})$. If the exact pose of the object is not known at the beginning of the query, $\state_i$, or it is hard to predict where the object ends up at the end of the query, $\state_{i+1}$, this is liable to cause grounding failure. Another example involves pushing an object on an unknown surface. This may result in a stochastic end pose of the object that is different from what was planned.

\begin{figure}
    \includegraphics[width=0.485\textwidth]
    {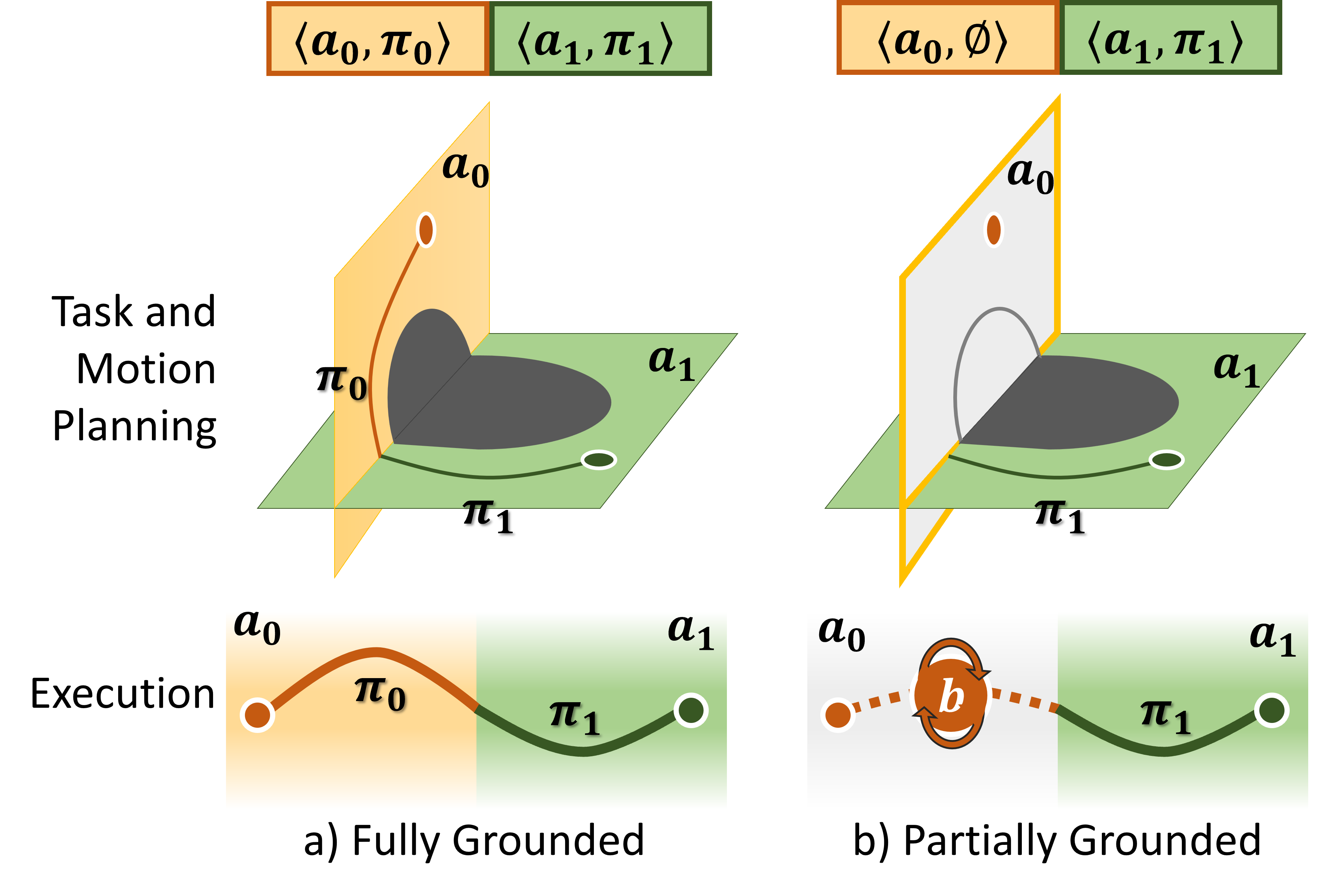}
    \vspace{-0.1in}
    \caption{The figure is a visual representation of grounding task and motion plans during planning and execution.
    The top row shows the task and motion plans.
    The middle visualizes the corresponding trajectories $\traj_i$ in each configuration space.
    The bottom shows the corresponding trajectories in the real-world during execution.
    The colored spaces (yellow and green) represent individual symbolic states.
    \textbf{a)} In the fully grounded case, the real world closely matches the planning model, which is known. The is typical to classical TAMP methods. 
    \textbf{b)} In the partially grounded case, 
    the light gray regions indicate that we do not have good enough knowledge to ground action $\action_0$.
    With a typical TAMP formulation, we cannot even plan the motion for $\action_0$ due to the gap. With our work, we propose to leave a gap in the task and motion plan ($\langle \action_0, \emptyset \rangle$, which is filled by a behavior $\behavior$ during execution (represented by the orange circle and dotted line).
    }
    \vspace{-0.15in}
    \label{fig:all-gaps}
\end{figure}

\noindent\textbf{Partial Grounding:}
A symbolic transition  $\langle \state_i, \action_i, \state_{i+1} \rangle$ in a task plan is partially grounded if either, \textbf{a)} a motion planning query cannot be constructed and computed, or \textbf{b)} the end world state of executing $\traj_i$ can differ substantially from the $\wstate_{i+1}$ that was planned. Such a grounding failure corresponds to a grounding \textit{gap} in the task and motion domain causing an empty trajectory ($\traj_i=\emptyset$) in a task and motion plan.

\noindent\textbf{Behavior:}
A behavior $\behavior$ is a closed-loop module that controls the robot joints and perception systems and is capable of reasoning during execution. Each behavior corresponds to some symbolic action and if successful can execute real-robot motions for a symbolic transition $\langle \state_i, \action_i, \state_{i+1} \rangle$. A successful behavior must satisfy: $\wstate_i\in \wmap(\state_i)$ and the end world state in execution should satisfy $\wstate_{i+1} \in \wmap(\state_{i+1})$, i.e., the grounding gap is bridged, and the symbolic pre-condition and end-effect of $\action_i$ are satisfied. A set of behaviors is $\Behaviors$.

The behavior can be thought of as a local policy designed for a corresponding action. The general framework uses whatever behaviors are available and can be attempted during execution. The implementation can take the form of handcrafted subroutines, local policy modules, or general architectures like BTs~\cite{colledanchise2016behavior}. They can also be learned controllers~\cite{wang2022hierarchical}.
Some design choices are shown in Sec~\ref{sec:real}.
Note that the design of such behaviors is usually domain-specific, and is not the focus of this paper.
The framework is formulated and presented to admit any of these design choices as long as the behavior is capable of using online information to compute and execute a motion for the action.
We focus on how we leverage such behaviors with powerful TAMP solvers to overcome the grounding gaps, as elaborated in Sec.~\ref{sec:method}. An example on how a behavior might fill in a gap is shown in Fig.~\ref{fig:all-gaps}b.

\noindent\textbf{Partially Grounded Task and Motion Plan:}
Given a task plan $(\action_{0}, ..., \action_{n})$, where each $\action_i$ transitions from symbolic state $\state_i$ to $\state_{i+1}$, and the task plan transitions from the initial state $\state_{0}$ 
to a goal state $\state_{n} \in \States_{goal}$, a partially grounded task and motion plan is $$ \tmeplan = (\langle \action_{0}, \traj_{0}\rangle, ..., \langle \action_{n}, \traj_{n}\rangle)$$ where actions that can be not grounded are allowed (i.e., $\traj_i = \emptyset$ is allowed).
In order to be executed, a partially grounded plan requires $$\forall \traj_i = \emptyset, \exists \behavior\in\Behaviors \text{ for } \action_i$$ where a behavior is available for each action that is not grounded.

\noindent\textbf{\fullname\text{} (\acronym):}
Given a task planning domain $\tpdomain$, a motion planning domain $\mpdomain$, a set of given behaviors $\Behaviors$, and a set of goal states $\States_{goal}$, \acronym is the problem of interleaved computation and execution of partially grounded task and motion plans ($\tmeplan$) consisting of a sequence of valid actions, motion, and closed-loop online behaviors that enable the robot to reach a goal state $\state \in \States_{goal}$ in the real world. 

\begin{table}[]
\caption{Glossary of Terms}
\vspace{-0.1in}
\renewcommand{\arraystretch}{1.2}
    \centering
    \begin{tabular}{p{0.4in}|p{2.7in}}
        \textbf{Terms} & \textbf{Definition} \\ \hline
        $\tpdomain$ & Task domain \\ \hline
        $\States$, $\state$ & Set of states and a state in task domain \\ \hline 
        $\Actions$, $\action$ & Set of actions and an action in task domain \\ \hline 
        $\Constraints$, $\constraint$ & Set of constraints and a constraint in task domain \\ \hline 
        $\mpdomain$ & Motion domain \\ \hline 
        $\cspace$, $\config$ & Configuration space and a configuration \\ \hline 
        $\traj$ & Motion plan in configuration space \\ \hline
        $\tmplan$ & Task \& motion plan \\ \hline 
        $\WStates$, $\wstate$ & World state space and a world state \\ \hline 
        $\wmap$ & Mapping from a symbolic state to set of world states\\ \hline 
        $\Behaviors, \behavior$ & A set of behaviors and a behavior \\ \hline 
        $\tmeplan$ & Partially grounded task \& motion plan \\ \hline 
    \end{tabular}
    \label{tab:glossary}
    \vspace{-0.1in}
\end{table}

\section{Method}
\label{sec:method}
We propose a framework for TAMPER capable of reasoning over partial groundings and real-world execution feedback to solve long-horizon tasks.
The framework consists of a planning layer and an execution layer (Fig.~\ref{fig:tamper-block}). In planning, we use an existing TAMP planner to plan as much as we can using the current, incomplete model of the world to discover motion constraints and find a partially grounded plan with gaps. During execution, we leverage 
online behaviors to fill in gaps in the partial plan and feed back failures from execution as new constraints, completing only upon real-world success.

\subsection{Foundations: Task and Motion Planning}

\revision{Consider a standard incremental constraint-based task and motion planner in Alg.~\ref{alg:tamp}, introduced in previous work~\cite{dantam2018incremental}
}
It assumes a perfect model of the world is given.
The input to the algorithm is the task planning domain, the initial state, the goal described as some symbolic expression, and the model of the world for motion planning.

\definecolor{primarycolor}{RGB}{33,49,77}   %
\definecolor{angrycolor}{RGB}{210,73,42}    %
\definecolor{lightcolor}{RGB}{146,162,189} %
\definecolor{bgfillcolor}{RGB}{206,213,221} %
\newcommand\mycommfont[1]{\ttfamily\textcolor{angrycolor}{#1}}
\newcommand\mykwfont[1]{\textbf{\textcolor{primarycolor}{#1}}}
\SetCommentSty{mycommfont}
\SetKwSty{mykwfont}
\newcommand\Break[0]{\KwSty{break}}
\newcommand{\compress}{\vspace{0.1in}}
\SetAlgoSkip{compress}

\setlength{\textfloatsep}{0pt}%
\begin{algorithm}[h]
  \caption{\textsc{TAMP}~\cite{dantam2018incremental}}
  \label{alg:tamp}
  \KwIn{$\tpdomain, \mpdomain, \state_{init}, \States_{goal}, world$}
  
  $\state \gets \state_{init};\ \Constraints \gets  \algsc{NoConstraint};\ \tmplan \gets \algsc{NoPlan};$

  \tcp*[h]{\footnotesize Planning Loop}
  
    \While{$\tmplan = \algsc{NoPlan}$}
    {
    
    \tcp*[h]{\footnotesize Task Planning}
  
        \If{$\tplan \gets \textsc{TP}(\state, \States_{goal}, \Constraints, \tpdomain)\ \algresult{fails}$} 
        {
            \label{alg:tamp:tp}
            \If{$\Constraints=\algsc{NoConstraint}$}{\algterm{return} \algsc{NoSolution};}
            $\Constraints\gets\algsc{NoConstraint};$ 
        }
        
       \tcp*[h]{\footnotesize Motion Planning}
  
        \ForEach{$\action_i \in \tplan$}{
            \If{$\traj_{i} \gets \textsc{MP}(\action_i, \mpdomain)\ \algresult{succeeds}$}
            {
            \label{alg:tamp:mp}
            $\tmplan \gets \tmplan \oplus  \langle\action_i, \traj_i\rangle;$
            }
            \Else
            {
                $\Constraints \gets \Constraints \cup \textsc{Constraint}(\action_i);$
                \label{alg:tamp:constraint}
                
                $\tmplan \gets \algsc{NoPlan};$
                \label{alg:tamp:reset}
                    
                \Break;
            }
        }
            
    }

    \tcp*[h]{\footnotesize Open-loop Execution}

    \ForEach{$\langle\action_{i}, \traj_{i}\rangle \in \tmplan$}{
        $\textsc{Execute}(\traj_{i})$
        \label{alg:tamp:execute}
    
    }
                
  \algterm{return} \algsc{Success};

\end{algorithm}
  \vspace{-0.1in}

An SMT-solver as the task planner supports the incremental-solving feature, leveraged by many existing planners~\cite{dantam2018incremental, pan2021multiple_manipulators, thomason2022tmit}. A candidate task plan is a sequence of symbolic actions
\revision{(Alg.~\ref{alg:tamp} line~\ref{alg:tamp:tp})}
which needs to be grounded.
The grounding of action $\action_i$ creates a motion planning query from the current world state $\wstate_i$, and computes a trajectory $\traj_i$ using a sampling-based motion planner
\revision{(Alg.~\ref{alg:tamp} line~\ref{alg:tamp:mp})}.
Motion planning can fail for some actions.
These are then added as a symbolic motion constraint 
\revision{(Alg.~\ref{alg:tamp} line~\ref{alg:tamp:constraint})
to the constraint stack of the SMT-based task planner, 
and a new task plan queried.}
\revision{The incremental solving feature 
maintains the guarantee of finding a feasible 
plan with the minimum number of actions.}
If the task planner cannot find any alternate plans, 
\revision{
it reports that the problem is either infeasible, or 
the constraints should be cleared to start over again with increased motion planning time budget
(Alg.~\ref{alg:tamp} line~\ref{alg:tamp:reset}).
When the planner successfully finds a task and motion plan $\tmplan$, it is typically sent to the robot for 
\textit{open-loop}
execution
(Alg.~\ref{alg:tamp} line~\ref{alg:tamp:execute}).
}
\cacceptrevision{This planner serves as a module in our proposed method, including necessary changes enabling it to work for grounding steps with knowledge gaps. The details are reflected in Alg.~\ref{alg:tamper} and explained in Sec.~\ref{sec:method:framework}.}

\subsection{\acronym Framework}
\label{sec:method:framework}

We now present our new algorithm (Alg.~\ref{alg:tamper}) for task and motion planning and execution reasoning (TAMPER).

\begin{wrapfigure}{r}{0.5\linewidth}
    \centering
    \vspace{-0.18in}
    \includegraphics[width=0.99\linewidth]{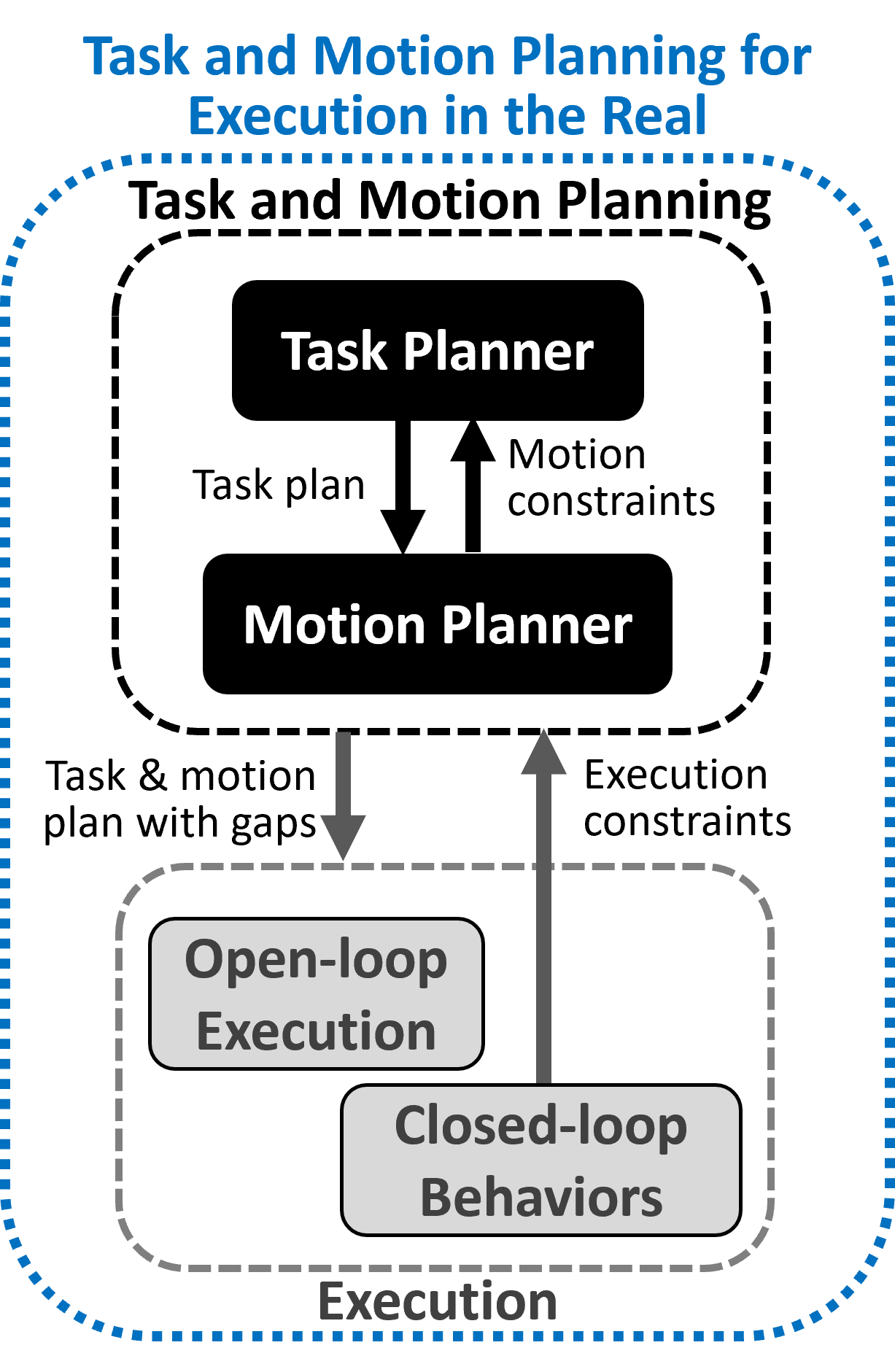}
    \vspace{-0.3in}
    \caption{The \acronym framework.}
    \vspace{-0.13in}
    \label{fig:tamper-block}
\end{wrapfigure}
We make the assumption that a set of behaviors corresponding to a subset of actions in the task domain is assumed to be available as input.
Hence, the input to Alg.~\ref{alg:tamper} is the task planning domain, the initial symbolic state $\state_{init}$, the set of goal states that can be encoded as a symbolic expression, the model of the world, and a set of pre-defined behaviors.
Our framework (Fig.~\ref{fig:tamper-block}) solves such a problem by interleaving planning and execution 
described as follows.

\definecolor{primarycolor}{RGB}{33,49,77}   %
\definecolor{angrycolor}{RGB}{210,73,42}    %
\definecolor{lightcolor}{RGB}{146,162,189} %
\definecolor{bgfillcolor}{RGB}{206,213,221} %
\SetCommentSty{mycommfont}
\SetKwSty{mykwfont}
\SetAlgoSkip{compress}

\setlength{\textfloatsep}{0pt}%
\begin{algorithm}[ht]
  \caption{\textsc{TAMPER}}
  \label{alg:tamper}
  \KwIn{$\tpdomain, \mpdomain, \state_{init}, \States_{goal}, \Behaviors, world$}
  
  $\state \gets \state_{init};\ \Constraints \gets  \algsc{NoConstraint};\ \tmeplan \gets \algsc{NoPlan};$
  \label{alg:tamper:state-init}

  \tcp*[h]{\footnotesize Planning and Execution Loop}
  
  \While{\algterm{not} $\textsc{IsSatisfied}(\state, \States_{goal})$}{
  
    \tcp*[h]{\footnotesize Partially-grounded TAMP}
    
    \While{$\tmeplan = \algsc{NoPlan}$}{
    \label{alg:tamper:partial-tamp}
    
    \tcp*[h]{\footnotesize Task Planning}
  
        \If{$\tplan \gets \textsc{TP}(\state, \States_{goal}, \Constraints, \tpdomain)\ \algresult{fails}$\label{alg:tamper:tp}} {
            \If{$\Constraints=\algsc{NoConstraint}$}{\algterm{return} \algsc{NoSolution};}
            $\Constraints\gets\algsc{NoConstraint}; 
            $\label{alg:tamper:reset}
        }
        
       \tcp*[h]{\footnotesize Grounding and Motion Planning}
  
        \ForEach{$\action_i \in \tplan$}{
            $\langle\action,\traj\rangle \gets \langle\emptyset,\emptyset\rangle;$
            
            \If{$\textsc{canGround}(\action_i,world)$}
            {
            \label{alg:tamper:canGround}
                \If{$\traj_{i} \gets \textsc{MP}(\action_i, \mpdomain)\ \algresult{succeeds}$}
                {
                \label{alg:tamper:ground}
                    $\langle\action,\traj\rangle \gets  \langle\action_i, \traj_{i}\rangle;$
                }
            }
            \ElseIf{$\textsc{hasBehavior}(\action_i, \Behaviors)$}
            {
            
                \tcp*[h]{\footnotesize Behaviors for Failed Grounding}
                
                \label{alg:tamper:hasBehavior}
                    $\langle\action,\traj\rangle \gets \langle\action_i, \emptyset\rangle;$
                    \label{alg:tamper:gap}
            }

            \If{$\langle \action,\traj \rangle = \langle \emptyset,\emptyset \rangle$}
            {
            $\Constraints \gets \Constraints \cup \textsc{Constraint}(\action_i);$
            \label{alg:tamper:constraint}
            
            $\tmeplan \gets \algsc{NoPlan};$
                
            \Break;
            }
            
            $\tmeplan \gets \tmeplan \oplus  \langle\action, \traj\rangle;$
            \label{alg:tamper:append}
        }
    }

    \tcp*[h]{\footnotesize Execution with Behaviors and Feedback}
    
    \ForEach{$\langle\action_{i}, \traj_{i}\rangle \in \tmeplan$}{
        \If{$\traj_{i} \neq \emptyset $
        }{$\textsc{Execute}(\traj_{i})$\label{alg:tamper:execute}}
        \ElseIf{$\behavior \gets \textsc{Behavior}(\action_i, \Behaviors)$ \algresult{fails} \label{alg:tamper:behavior} \algterm{or} $   \tmeplan \gets \textsc{Repair}(\tmeplan)$ \algresult{fails} }{
                $\Constraints \gets \Constraints\cup\textsc{Constraint}(\action_i);$
                \label{alg:tamper:constraint3}

                \Break;
        }
    
    }
    $\state \gets \textsc{getCurrentState}(world);$
                \label{alg:tamper:infer-state}            
    $\tmeplan \gets \algsc{NoPlan};$
                
  }
  \algterm{return} \algsc{Success};
\end{algorithm}

\subsubsection{Planning}
The task planner computes a candidate task plan from $\state_{init}$ (Alg.~\ref{alg:tamper} line~\ref{alg:tamper:tp}). For each action in the plan, we first check if we can ground it 
\revision{using \textsc{canGround} (Alg.~\ref{alg:tamper} line~\ref{alg:tamper:canGround}), which}
is a typical TAMP process that creates motion planning queries. In manipulation problems, it usually involves first computing grasp poses of the objects, or placing poses of the objects, and then computing the robot configurations achieving these poses as the motion planning goal. Some examples of the grounding process of the actions that we use in the experiments are introduced \revision{in details in Sec.~\ref{sec:real}.}
When \textsc{canGround} outputs false, it is considered that this action cannot be grounded. For example, if the action is to pick up some object from the table, but the object is occluded from the sensor and its pose is unknown, \textsc{canGround} should output false.
Note that here we allow planning with incomplete models, so we consider that \textsc{canGround} encodes the confidence on the model of the world. For instance, if the friction of a surface is inaccurately modeled, the consequence of a \textit{push} might not be reliably planned.

If \textsc{canGround} decides that the action can be grounded, the motion planning query is sent to a sampling-based motion planner (Alg.~\ref{alg:tamper} line~\ref{alg:tamper:ground}).
If the motion planner fails to compute a trajectory, similar to \tamp, the failure is encoded as a symbolic constraint (Alg.~\ref{alg:tamper}, line~\ref{alg:tamper:constraint}), and pushed to the constraint stack to compute a new alternate task plan that does not include the particular action.

If \textsc{canGround} determines an action $\action_i$ cannot be grounded with the current model of the world, we check if any one of the given behaviors is applicable for $\action_i$ (Alg.~\ref{alg:tamper} line~\ref{alg:tamper:hasBehavior}). If a behavior is available, we add an empty trajectory to the task and motion plan (Alg.~\ref{alg:tamper} line~\ref{alg:tamper:gap}), indicating that the grounding of this action is deferred to the execution level.
To continue grounding the next action $\action_{i+1}$, we \revision{use an 
optimistic 
} robot configuration to construct a \revision{safe} world state heuristically in the planner as the start state of the next action.
\revision{In our implementation, we design a repair strategy to rejoin to any of the configurations along $\traj_{i+1}$ during execution (see the Sec.~\ref{sec:experiments:framework} for details).}
If there is no associated behavior, it is considered a grounding failure, and also encoded as a symbolic constraint (Alg.~\ref{alg:tamper} line~\ref{alg:tamper:constraint}), to find a new task plan. The output of the planning part is a partially-grounded task and motion plan $\tmeplan$, which allows gaps (empty trajectories $\traj_i = \emptyset$) to be grounded by the behaviors in execution.
Note that if the input $\Behaviors = \emptyset$, i.e., no behaviors are given, our planner reduces to a typical TAMP planner~\cite{dantam2018incremental}.

\subsubsection{Execution}
After a partially-grounded task and motion plan is computed, the execution layer executes the trajectories in sequence (Alg.~\ref{alg:tamper} line~\ref{alg:tamper:execute}). If $\traj_i$ is empty, we execute the associated behavior for this action $\action_i$. A behavior module may involve online sensing, planning, and execution.
We assume a set of behaviors $\Behaviors$ are available as input. These can be simple sequential steps, BTs~\cite{colledanchise2016behavior} designed by human experts, local policies that can be learned controllers~\cite{wang2022hierarchical}, etc. The details of the behaviors that we use are discussed in Sec.~\ref{sec:experiments:framework}.

When the behavior succeeds, online trajectory repair is performed to join to the next pre-computed trajectory $\traj_{i+1}$ using \textsc{Repair} (Alg.~\ref{alg:tamper} line~\ref{alg:tamper:behavior}).
Different repairing strategies exist for various replanning frameworks, and
any strategy can be used here, as long as it joins to one of the configurations $\traj_{i+1}(\tau), \tau\in[0,\trajend]$. \textsc{Repair} returns the updated $\traj_{i+1}$ to be executed next. If $\traj_{i+1}$ was also empty, it will be handled by the next behavior, and \textsc{Repair} does not need to update it.
\revision{

In this work, we do \textit{not} assume that behavior execution and the \textsc{Repair} module always succeed.
The behavior execution for \cacceptrevision{$\action_{i}$} might fail because the necessary information to allow its successful execution may still not have been discovered after executing pre-computed trajectories or behaviors of all the actions from $\action_0$ to \cacceptrevision{$\action_{i-1}$. This suggests that we might not be able to ground $\action_{i}$ from the current state, requiring to recompute a new task and motion plan.}
The $\textsc{Repair}$ following executing the behavior of $\action_{i}$ may fail due to not finding a motion plan to the initial state of $\traj_{i+1}$ (or other waypoints in $\traj_{i+1}$, defined by the actual implementation of $\textsc{Repair}$ detailed in Sec.~\ref{sec:experiments:framework}) within the time limit.
\cacceptrevision{This indicates that the execution of the behavior of $\action_{i}$ succeeded in the real world, but the resulting state of the world might make the next action $\action_{i+1}$ not feasible.}
}
\revision{When any of such execution failures happen, similar in spirit to incremental \tamp,}
we can encode them as symbolic constraints (Alg.~\ref{alg:tamper} line~\ref{alg:tamper:constraint3}) for replanning \revision{--- as opposed to typical \tamp which replans within the planning loop, we are \textit{replanning within a planning and execution loop}}. We infer the current state by combining the previous knowledge as well as new information from sensing (Alg.~\ref{alg:tamper} line~\ref{alg:tamper:infer-state}). 
Then, we return to the planner layer, and replan from the current state (Alg.~\ref{alg:tamper} line~\ref{alg:tamper:tp}). Replanning uses all the constraints we have discovered from the planning and execution so far, improving the efficiency of solution discovery and execution.

It is worth noting that in the algorithm we use a symbolic state $\state$ to track the world state, only to keep it concise. All the state variables (e.g., object poses) are essentially kept track of throughout the framework for both planning and execution.

\textit{Note on Planning vs. Execution Trade-off: }
For a typical TAMP planner~\cite{dantam2018incremental}, it always calls the task planner to find an alternate candidate plan if it fails to ground an action.
In our proposed algorithm, we prioritize using given behaviors when an action cannot be grounded, rather than relying on the task planner to compute another candidate plan (Alg.~\ref{alg:tamper} line~\ref{alg:tamper:hasBehavior}).
\revision{
Similar to \tamp methods~\cite{dantam2018incremental, garrett2020pddlstream, shah2020anytime}, a model of the world (domain) is available to plan over within \tamp as well as in the behaviors. If the execution is inconsistent with the domain and errors in the pipeline induce unrecoverable dead-end states (where the goal becomes unreachable), such execution will trigger task failures. Note the proposed method still operates within an available and accurate task domain, while expanding the class of problems we can address using an enhanced paradigm built on classical \tamp.
}
This trade-off is discussed in details in Sec.~\ref{sec:discussion}.
\revision{A discussion on theoretical guarantees such as probabilistic completeness in the domain combining planning and execution is also presented in Sec.~\ref{sec:discussion}.}

\vspace{-0.1in}

\subsection{\revision{Necessary Assumptions and Modules}}
\revision{We present the considerations that our proposed framework relies on to extend the applicability of traditional \tamp (Alg.~\ref{alg:tamp}), where \tamp typically assumes open-loop execution~\cite{dantam2018incremental,garrett2020pddlstream,srivastava2014combined}.
Notably, many of these assumptions are not unique to the proposed approach and are known to come up when typical \tamp methods such as Alg.~\ref{alg:tamp} have been implemented in practice~\cite{garrett2021review, Toussaint2022sequence}.
We argue that these assumptions are reasonable and necessary because without our framework, the class of problems we study in this paper either cannot be addressed at all, lead to infeasibility, or yield poor empirical performance (as demonstrated in Sec.~\ref{sec:experiments}).
}
\begin{itemize}[leftmargin=*]
    \item We assume the set of behaviors is provided as input accompanying the specific problem domain.
\revision{How the behaviors can be more efficiently designed or learnt based on the execution results of the proposed framework is left for future work with some discussion in Sec.~\ref{sec:discussion}.}

    \item \revision{Each symbolic action is associated with provided domain knowledge on an evaluation of the $\textsc{canGround}$ function. 
    This knowledge is encoded as a part of $\textsc{canGround}$.
    The availability (and success) of such a module is a common assumption in typical TAMP methods, which rely on robust or well-modeled perception and execution modules. It is either the case that the world state is known and actions are assumed to be deterministic~\cite{dantam2018incremental, srivastava2014combined, garrett2020pddlstream}, or it is assumed the robot knows that stochasticity exists in the domain and the TAMP problems are solved based on some {known models of uncertainty}~\cite{shah2020anytime, pan2022failure, garrett2020online}.}
    \revision{In this work, we can address problems where the perception fails or when models of uncertainty are unknown or inaccurate. This is possible using execution-level behaviors that are available and performant across these grounding failures. The flexibility of the proposed framework comes at the behest of having access to the $\textsc{canGround}$ function that \textit{knows what is not known} or cannot be grounded. Typically such information might already be readily available in \tamp implementations --- consider the case where a pose estimator is invoked for an object in the planning domain and reports no detection, or the case where no robust simulator exists for grounding a symbolic action. Such failures occur but are not addressable by a standard \tamp framework.   
    }

    \item \revision{The framework needs a $\textsc{REPAIR}$ module during execution to be deployed after the completion of a behavior to rejoin the precomputed task and motion plan. Our experiments show that this module typically managed to use motion planning to \textit{rejoin} the subsequent trajectory. When it fails, the framework reverts to \tamp replanning \cite{curtis2022M0M} within the high-level loop of Alg.~\ref{alg:tamper}}. 

\end{itemize}

\section{Real-World Benchmarks and Design Choices}
\label{sec:real}
\begin{figure}[t]
    \centering
    \includegraphics[width=0.49\textwidth]{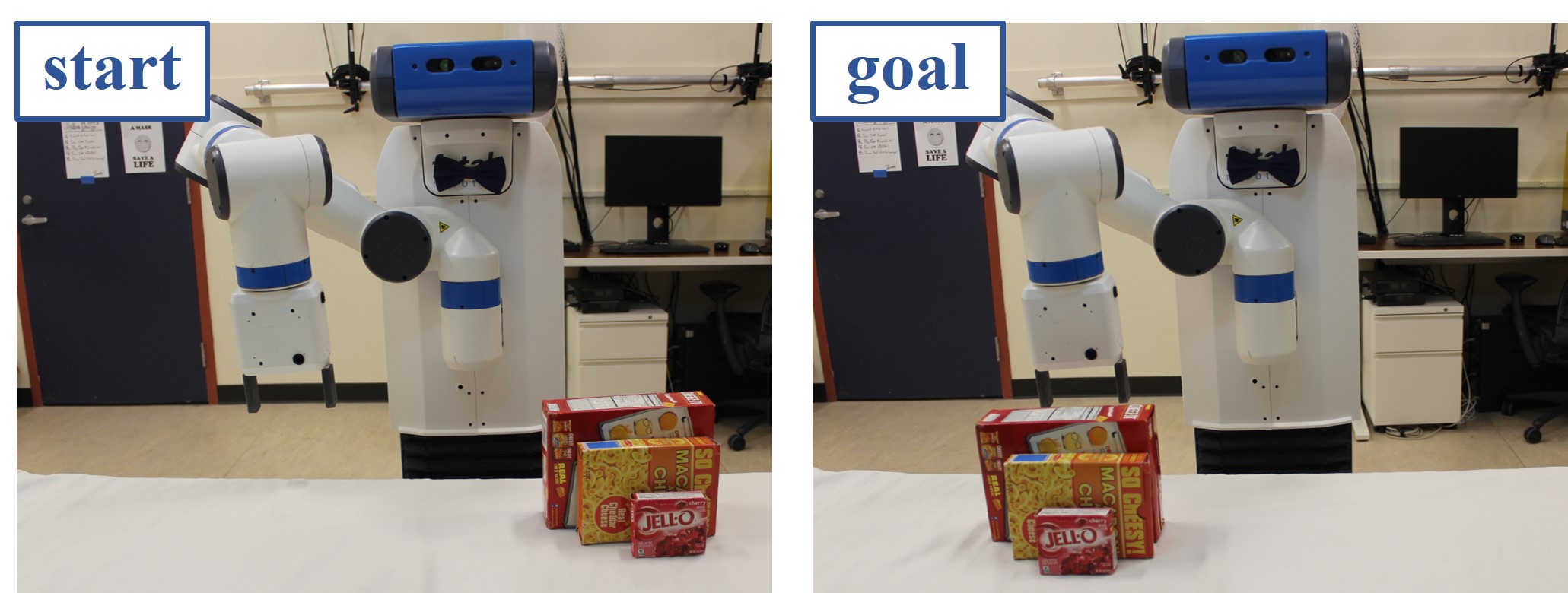}
    \vspace{-0.25in}
    \caption{\textbf{Horizontal Stacking} Benchmark: The task is to move the objects to the goal positions. The initial poses of the objects are designed to have some occlusion. Strict ordering constraints exist because smaller objects cannot be grasped in the presence of a nearby larger object.}
    \vspace{0.05in}
    \label{fig:h3-setup}
\end{figure}

We focus on the evaluation of our method on real-world scenarios. In this section we introduce two scenarios and our necessary design choices dictated by the target scenarios.

\subsection{Benchmarks}
Our real-world experiments are performed on a Fetch robot with an RGBD sensor mounted on its head and objects from the YCB dataset~\cite{calli2015ycb} and the HOPE dataset~\cite{tyree2022hope}. Vicon Cameras are used for pose estimation of the obstacles such as the table and the drawer. A deep learning-based perception module (DOPE~\cite{Tremblay2018DeepOP}) is used to detect the objects. The task planner is implemented using Z3 planner~\cite{moura2008z3}, and the motion planner is RRT-Connect~\cite{kuffner2000rrt} via Robowflex~\cite{kingston2022robowflex}.

\subsubsection{Horizontal Stacking}
As shown in Fig.~\ref{fig:h3-setup}, there are three objects on the table \revision{with known geometric models}.
The initial object poses are designed to create occlusions for the camera. The goal is to move all the objects to the goal positions, which are explicitly specified to allow only one order of placement. If a large object is placed to its goal position first, the geometry blocks the robot arm from placing a smaller object to its goal position. The available actions are \pick\text{} and \place. 

\subsubsection{Kitchen Arrangement}
\begin{figure}[t]
    \centering
    \includegraphics[width=0.49\textwidth]{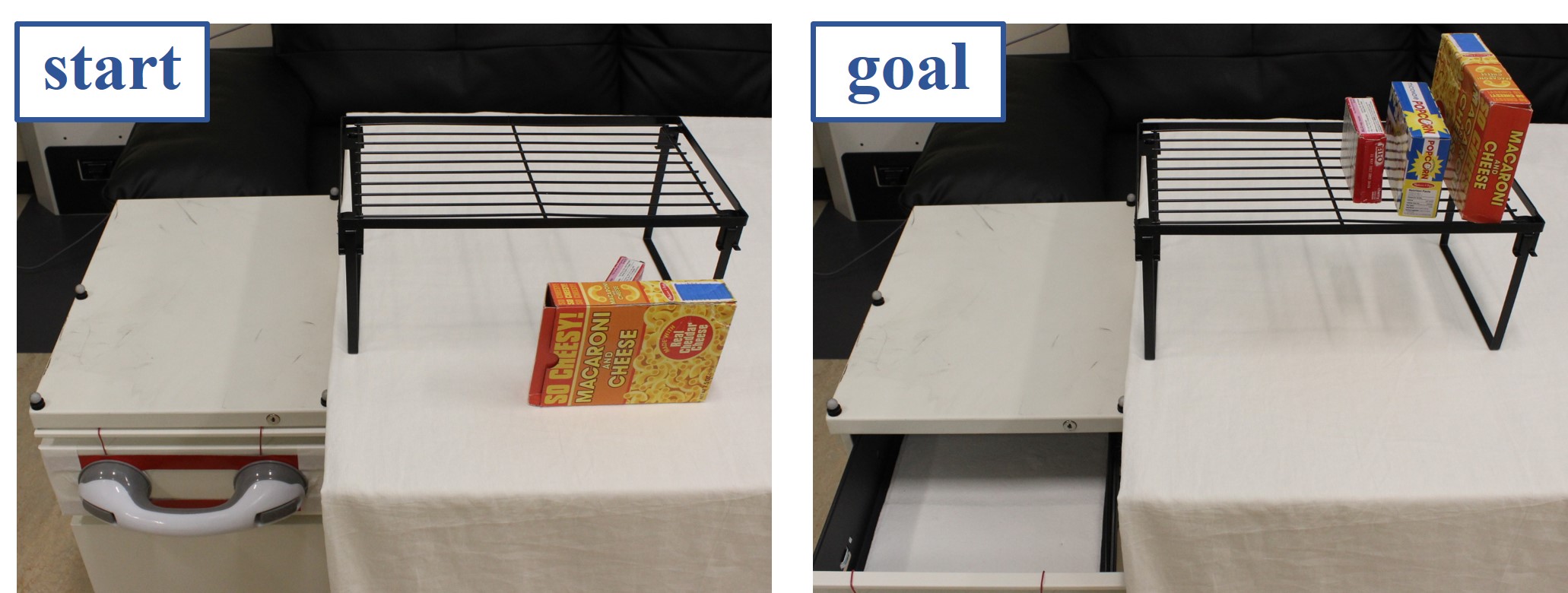}
    \vspace{-0.25in}
    \caption{\textbf{Kitchen Arrangement} Benchmark: Move the objects to the goal positions. The starting position of one of the objects is inside the drawer. In this case, there is both occlusion and uncertainty in execution outcomes.}
    \vspace{0.05in}
    \label{fig:ds-setup}
\end{figure}

We add a drawer and a shelf to the scene (as in Fig.~\ref{fig:ds-setup}). Initially, one object is in a closed drawer, and two objects are on the table. When the small object is on the shelf, it can only be picked from its side. 
When the large object is close to the shelf, it cannot be picked from any direction. To address this challenge we define a \pushpick\text{} action. Grounding it involves first computing the motion to push an object away with a distance, and then computing the motion of picking.
We also define an \open\text{} action that opens a drawer, where grounding requires computing trajectories to approach and pull the drawer handle.
The goal is similar to the Horizontal Stacking, but on the shelf.
The available actions are \pick\text{}, \place\text{}, \open\text{} and \pushpick.

\subsection{Framework Design}
\label{sec:experiments:framework}
Here, we particularly focus on how we design the execution layer of the framework. Note that the following design choices are modular and can be replaced by other options. Such design efforts help
to address realistic problems.

\subsubsection{Symbolic Encoding}
Some previous works assume a discretization over the workspace into grid locations~\cite{dantam2018incremental, pan2021multiple_manipulators}, where each location is associated with a different symbol, which makes the state space unnecessarily large. We choose to only explicitly encode the regions/locations of interest into different symbols,
\revision{which is a common choice in TAMP literature~\cite{srivastava2014combined, shah2020anytime, garrett2020pddlstream}.}
For a large support surface like the table, we do not discretize it but instead only assign a single symbol to it.
\revision{Such an encoding is also helpful to gain efficiency dealing with partial models. For example, even the pose of an object is unknown, its symbolic state can be captured by a single predicate \textsc{OnTable()}.
}
We encode the essential manipulations such as \texttt{Pick}, \texttt{Place}, \texttt{Push}, and \texttt{OpenDrawer} as symbolic actions.
Sensing is not explicitly encoded as an action, even though it is critical for the benchmark problems that have partial groundings due to occlusions. Instead, sensing is performed online in the behaviors, which avoids inflating the state and action space.
Our framework balances the trade-off of designing a more complicated abstraction model versus designing a more powerful local behavior. Which one is better depends on the actual problem to be solved.

\subsubsection{Occlusion Model}
Since we purely rely on point cloud sensors to get geometric information of the real world, we address having access to partial observations due to occlusions, etc. To perform motion planning in such real-world manipulation problems, naive strategies can either assume there is nothing to avoid in the occluded area (most optimistic), or to perform collision-checking assuming the occluded area is fully occupied by something (most conservative). 

\begin{figure}[t]
    \centering
    \includegraphics[width=0.4\textwidth]{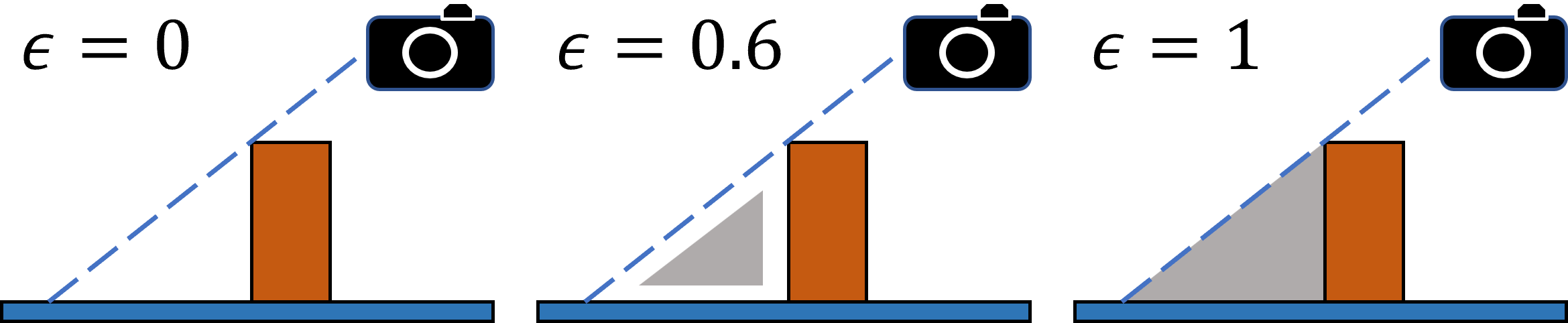}
    \vspace{0.02in}\hrule
    \includegraphics[width=0.4\textwidth]{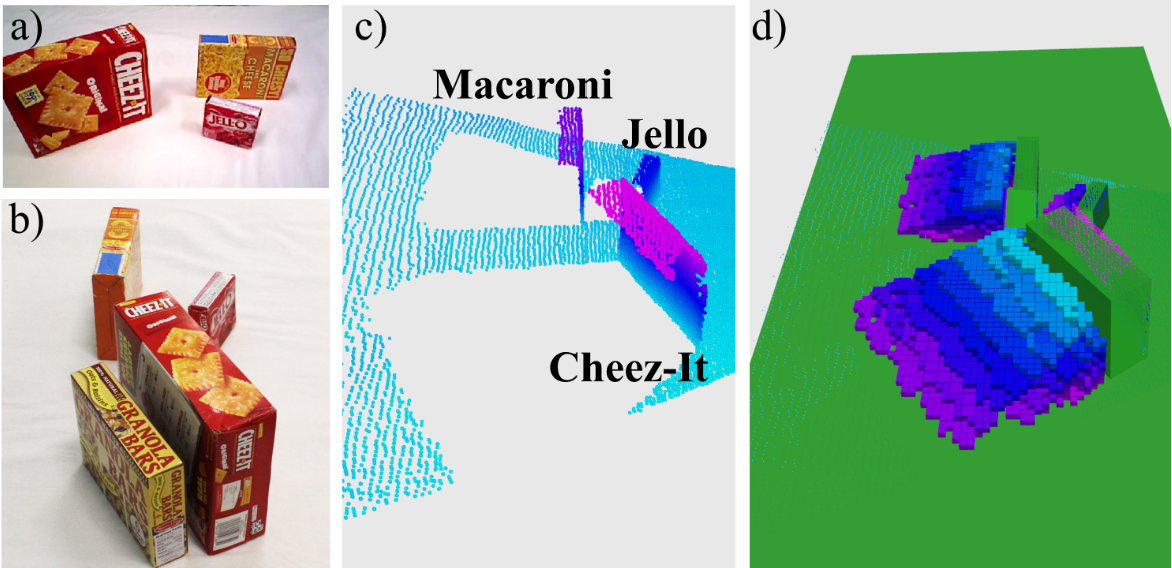}
    \caption{(\textbf{Top}) Our parameterized occlusion model when it is (\textbf{top-left}) most optimistic, with $\occl=0$, (\textbf{top-middle}) parameterized by $\occl\in[0,1]$, and (\textbf{top-right}) the most conservative  $\occl=1$.
    \textbf{(Bottom)} shows the real RGBD sensor. 
    \textbf{a)} RGB camera input.
    \textbf{b)} The side view of the scene (not available to robot).
    \textbf{c)} Point cloud detected by the camera \revision{(rotated to illustrate the occlusion)}.
    \textbf{d)} An example of our occlusion model shown in octomap with $\occl=1$.}
    \vspace{0.05in}
    \label{fig:occlusion_rviz}
\end{figure}

In our experiments, we use a parametric model of the occlusion inspired by~\cite{shimanuki2021shadow}. Given an input point cloud $\cloud$, we perform ray-casting of the points from the camera viewpoint to get a new cloud $\cloud_r$ where the occluded parts are also filled with points. Then, we crop out the points that correspond to the known objects (the objects that are recognized by the perception modules), and get the clusters that correspond to each object $\obj \in \objs$. Then, the occluded geometry caused by an object $\obj_1$ can be represented by the set of clusters $\{\cloud_{\obj_1}^1, ..., \cloud_{\obj_1}^m\}$. We compute the centroid of each cluster, and shrink the cluster with ratio $\occl_{\obj_1}\in [0,1]$. When the ratio $\occl_{\obj_1}=0$, it is the most optimistic representation, as it means there is nothing in the occluded area. When $\occl_{\obj_1}=1$, it is the most conservative representation, assuming the area is fully occupied. 
We use the same $\occl=0.6$ for all occlusions. Fig.~\ref{fig:occlusion_rviz} (top) shows the side view of our occlusion model with three different $\occl$. Fig.~\ref{fig:occlusion_rviz} (bottom) shows a visualization of our occlusion model in occupancy grids~\cite{hornung13octomap}.
Such an occlusion model is used in both the planning and 
the execution layer.

\subsubsection{Grounding Modules}
There are often cases that a task plan requires placing objects at intermediate locations. Choosing the placement poses effectively and feasibly is important in highly symbolically constrained problems. We encode the whole table surface as a region with a single symbol associated. Therefore, we need a placement location sampler to ground the \texttt{Place} action.
We perform uniform sampling within the robot's reachability area with certain constraints that take into account our occlusion model---avoiding intersections between the placement pose's occlusion model and the point cloud's occlusion model. 
To get configuration space goals, we sample valid inverse kinematic (IK) solutions, where we use the standard IK solver in MoveIt~\cite{sucan2013moveit}. The grasping problem is not the focus of the work. Where the object mesh is available, predefined axis-aligned grasps of objects are used.
\revision{When any of grounding modules cannot find a feasible value with a given time limit, it is considered as grounding failure and should be encoded as a symbolic constraint.}

\subsubsection{Behaviors}
All of the following behaviors are expert-designed, but they can also be learned policies,
etc.
\revision{Each behavior is executed step by step. If the execution of any step fails, it stops execution, and the failure is encoded as a constraint to trigger task-level replanning (Alg.~\ref{alg:tamper} line~\ref{alg:tamper:constraint3}).}

We use the following behavior for \texttt{Pick} in all benchmarks.

\begin{mybox}{\textbf{\texttt{Pick} Behavior\vspace{-0.08in}}}

    \tcblower

    \indent\textbf{1.} Invoke DOPE~\cite{Tremblay2018DeepOP} to detect the pose of the object.

    \indent\textbf{2.} If the object that we want to pick is detected compute a grasp, otherwise return failure.
    
    \indent\textbf{3.} Perform motion planning online and proceed to execution to grasp object.
\end{mybox}

We use the following behavior for the \texttt{Push-Pick} action in the \textit{Kitchen Arrangement} benchmark, which first pushes the object and then picks it up.

\begin{mybox}{\textbf{\texttt{Push-Pick} Behavior\vspace{-0.08in}}}

    \tcblower
    
    \indent\textbf{1.} Perform online constrained motion planning and execution to horizontally push the object along one of its axes with a fixed distance.

    \indent\textbf{2.} Use DOPE to detect the pose of the pushed object.

    \indent\textbf{3.} Compute axis-aligned grasp from its updated pose.

    \indent\textbf{4.} Perform motion planning online and proceed to execution to grasp object.
\end{mybox}

For the behavior of \texttt{OpenDrawer}, we design a sequence of key configurations parameterized by the pose of the drawer and plan and execute it online.
The behaviors can be implemented in several ways. For the above behaviors, we use sequential scripts. We use Behavior Trees~\cite{colledanchise2016behavior} to implement the behavior in the demo described in Sec.~\ref{sec:experiments:demo}. If the behavior fails, a constraint is fed back to the framework to compute a new partially grounded task and motion plan (Alg.~\ref{alg:tamper} line~\ref{alg:tamper:tp}).

\subsubsection{Repairing Strategy}
\label{sec:real:repair}
As discussed in Section~\ref{sec:method}, after the execution of a behavior $\behavior(\action_i)$ of action $\action_i$, we try to join to the pre-computed trajectory $\traj_{i+1}$ of the next action.
Any repairing strategy~\cite{koenig2002d} can be used as long as it plans for a path to one of the configurations in $\traj_{i+1}$, i.e., $\traj_{i+1}(\tau), \tau\in[0,\trajend]$.
\revision{Note that back in the task and motion planning stage, an initial state must be constructed to plan for $\traj_{i+1}$. Typically, any feasible state can be used since we only need to rejoin to one of the configurations in $\traj_{i+1}$, not necessarily only to $\traj_{i+1}(0)$. We choose to build the state from a robot configuration where the links are away from the domain-specified task space region, which is effective in practice. Fig.~\ref{fig:ds_pane}a shows an instance where the arm is raised away from the drawer and the table.
}

Our \textsc{Repair} strategy is to always try to join the first configuration $\traj_{i+1}(0)$. If this fails, then plan to rejoin $\traj_{i+1}(\trajend)$. It is possible that the geometries of the world change sufficiently and repair fails (or subsequent trajectories become infeasible). Failure here triggers replanning with the updated state.

\section{Results}
\label{sec:experiments}

In this section we report our empirical performance using data which has been exclusively collected from real-robot experiments.
More than 40 real-world physical trials have been used to compare our method with a replanning baseline.
It has to be stressed that the each experiment involves many-step planning and execution attempts, as well as multiple replanning tries upon sensed environment updates. 
Real-robot trials prove the most accurate, albeit more challenging and time-consuming, reflection of performance, where we demonstrate
clear performance benefits of our approach.

To promote replicability in other real-world setups as well as simulated ones, we share the experiment setup information from our benchmark trials~\footnote{Videos of all experiments are at: \url{https://kavrakilab.org/2023-pan-tamper/}}, including object poses, point clouds, RGB images, and domain models.
This is included in an open-source dataset directed towards real-robot TAMP applications, which we believe is one of the first of its kind.

\subsection{Evaluated Methods}

\subsubsection{\textbf{(Base)} Closed-loop TAMP Algorithm}
This baseline is implemented based on the method in \cite{curtis2022M0M}. The overall goal is given, which may involve objects that are not observed yet. It extracts a subgoal that only involves the observed objects, and greedily plans for this subgoal. Each time a TAMP plan is executed, it calls the perception module again, and plans for a new subgoal given new observations, until the task is completed. To ensure a fair comparison the baseline shares the same modules as the proposed method wherever it is used, like the task planner, motion planner, sampler, etc.

\subsubsection{\textbf{(Ours)} \acronym}
The proposed framework and design choices described in Sec.~\ref{sec:method} and \ref{sec:real}.

\subsection{Benchmark Results}
\begin{figure*}[ht!]
    \centering
    \vspace{-0.15in}
    \includegraphics[width=0.963\textwidth]{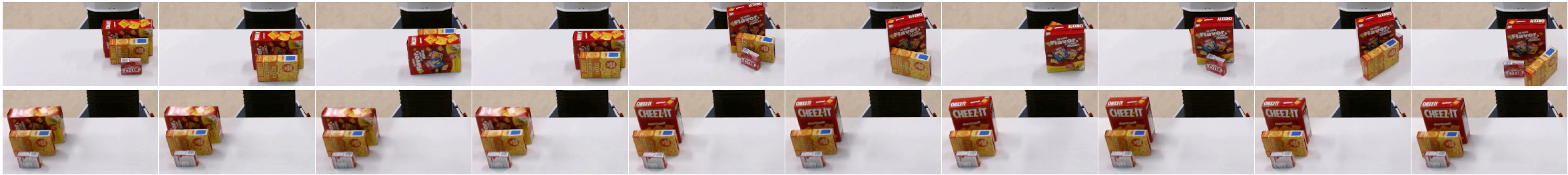}
    \vspace{-0.1in}
    \caption{The 10 problems in Horizontal Stacking. Each column shows the start (top) and the end (bottom) of \acronym's execution in each problem.}
    \label{fig:h3_startgoals}
\end{figure*}

Fig.~\ref{fig:h3_startgoals} summarizes the start and end of the 10 problems in Horizontal Stacking, and Table~\ref{fig:table1} shows the planning time and real-world execution time.
\revision{The first row (Plan-BE) shows how much time is spent in task and motion planning prior to the execution stage (i.e., the planning loop presented in Alg.~\ref{alg:tamper} from line~\ref{alg:tamper:partial-tamp} to~\ref{alg:tamper:append} encountered the first time). The second row shows how much time is devoted to planning after execution starts. It consists of both a) task and motion replanning, if triggered, and b) motion planning within the behaviors and the \textsc{Repair} module (Alg~\ref{alg:tamper} line~\ref{alg:tamper:behavior}).
The third row shows how much time is devoted to sensing modules after the execution begins. The fourth row exhibits the total time spent on executing trajectories (both the pre-computed ones and the online-computed ones in the behaviors) on the robot controllers. The fifth row is the sum of the numbers shown in the four rows above. The last two rows show the number of object interactions during solving the problem instance and whether it successfully accomplished the goal.
}
In this problem, the placement of the objects introduces ordering constraints since the robot cannot interact with the smaller objects while the larger objects are nearby. The relative size and position of the objects also introduce gaps due to occlusion.
\revision{In the 10 scenarios, the proposed method executed 9.2 \cacceptrevision{($\pm$ 0.98 one standard deviation)}  symbolic actions on average with 1.1 \cacceptrevision{($\pm$ 0.3)} behaviors  to complete the task.}
For all the problem scenarios, the proposed method takes much less computation during execution, and has fewer object interactions, with the overhead of slightly higher planning time before the execution starts.
\revision{Counting sensing, planning and execution,}
the total time to finish each task is much less compared to the baseline.

This demonstrates our advantage of leveraging TAMP with even an incomplete model of the real world, which discovers motion constraints during planning instead of executing expensive real-world motions to find the same set of constraints.

\begin{wrapfigure}{r}{0.5\linewidth}
    \centering
    \vspace{-0.1in}
    \includegraphics[trim={7.2cm 0cm 0cm 0cm},clip,width=0.99\linewidth]{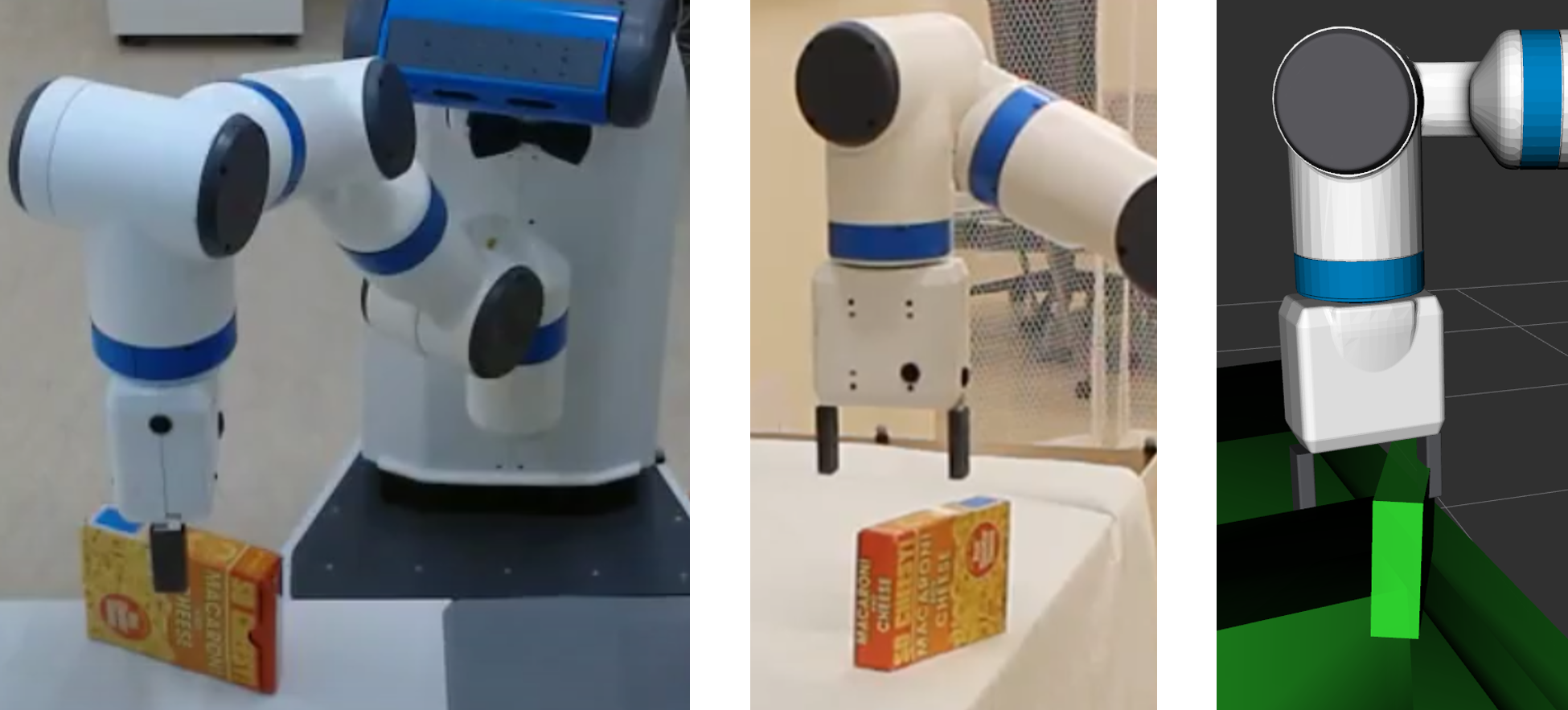}
    \vspace{-0.25in}
    \caption{An example of failure the baseline encounters with the Kitchen Arrangement benchmark. For safety, this trajectory was not executed.
    \textbf{Left}: The moment before the failure, side view.
    \textbf{Right}: The catastrophic failure of gripper-object collision that would have happened, replicated in RViz from 
    the plan and object pose detection.
    }
    \vspace{-0.15in}
    \label{fig:catastrophic}
\end{wrapfigure}

\begin{figure*}[ht!]
    \centering
    \vspace{-0.15in}
    \includegraphics[width=0.965\textwidth]{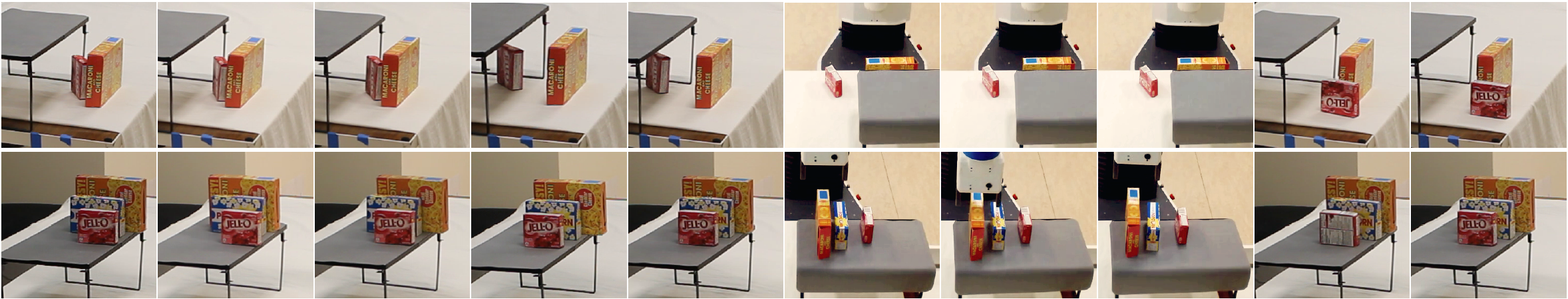}
    \vspace{-0.1in}
    \caption{The 10 problems in Kitchen Arrangement. Each column shows the start (top) and the end (bottom) of the execution of \acronym in each problem. Since the execution outcome of the push is stochastic, we run each scenario multiple times. From left to right, columns 1-3, 4-5, 6-8, and 9-10 show four different starting poses with either 3 runs or 2 runs each. Note that in all scenarios, one object is initially inside the drawer.}
    \label{fig:ds_startgoals}
    \vspace{-0.15in}
\end{figure*}

The start and end states reached by the proposed method over 10 problems in Kitchen Arrangement are shown in Fig.~\ref{fig:ds_startgoals} and the performance is recorded in Table~\ref{fig:table2}. 
Here, one of the object poses is unknown because it is inside the drawer, creating a gap. Another object is close to the edge of the shelf making it not possible to pick it directly.
A \pushpick\textbf{} action is also made available, that first pushes the object away from the shelf and then attempts to pick it up. Note that the baseline encounters catastrophic failure for all the scenarios. The reason is that the end state of pushing the object cannot be simulated perfectly in the planning stage. Any gaps here introduce undesirable interactions along subsequent actions.
Since the baseline only executes the computed motion plans sequentially, when the pose of the object is different than what was computed, the trajectory may be in collision (see Fig.~\ref{fig:catastrophic}). 

The proposed method leverages an online \pushpick\textbf{} behavior to achieve robustness of the execution of this action.
The proposed method succeeds in all the problems.
\revision{In the 10 scenarios, the proposed method executed on average 8.4 \cacceptrevision{($\pm$ 0.91)} symbolic actions with 2.3 \cacceptrevision{($\pm$ 0.46)} behaviors.}
\textit{Note: } The initial poses of the objects (Fig~\ref{fig:h3_startgoals} and \ref{fig:ds_startgoals}) are manually designed to create challenging task-motion interactions and gaps.
To ensure fair comparisons the poses are recorded and if the same setup is evaluated again, the poses are manually lined up by overlaying the point cloud and the recorded poses. This process is similar to the replication steps alluded to in the dataset discussion.

Example real-robot runthroughs of our proposed method on the Horizontal Stacking and the Kitchen Arrangement benchmarks are shown in Fig.~\ref{fig:h3_pane} and \ref{fig:ds_pane}.

\subsection{Grocery Demo}
\label{sec:experiments:demo}
\begin{wrapfigure}{r}{0.5\linewidth}
    \centering
    \vspace{-0.4in}
    \includegraphics[trim={0cm 0cm 0cm 0cm},clip,width=0.99\linewidth]{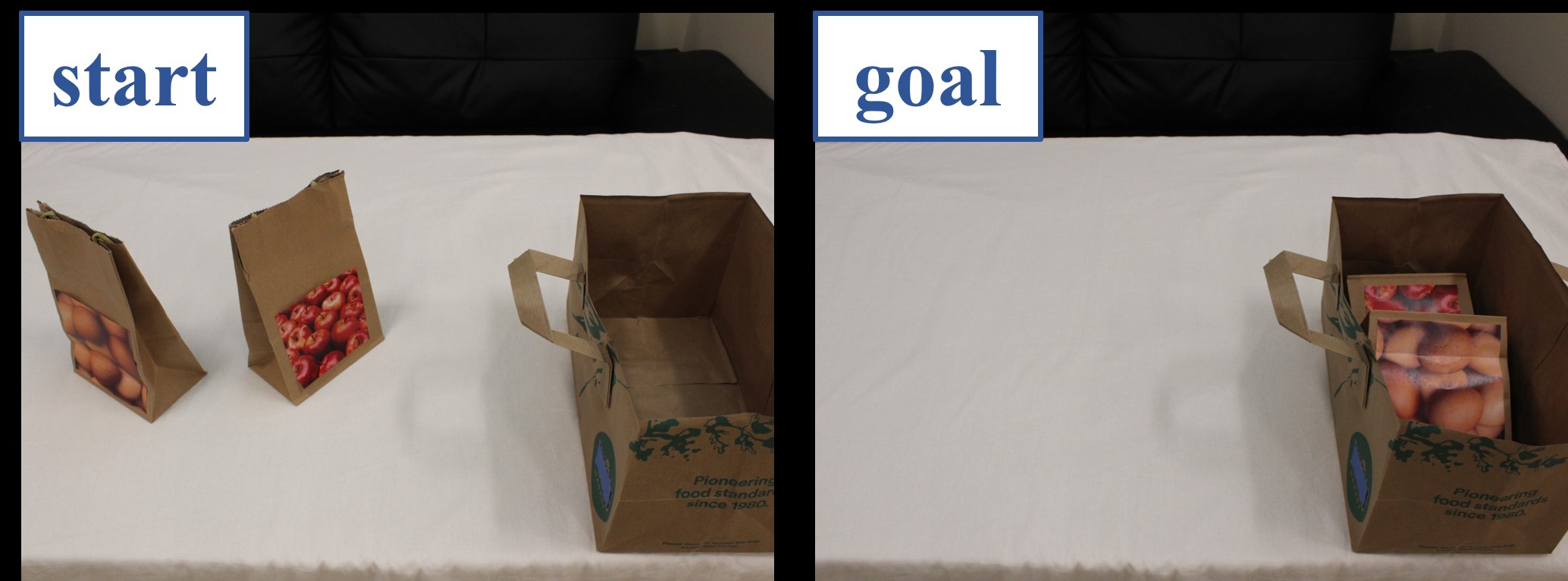}
    \vspace{-0.2in}
    \caption{The robot doing groceries. The objects must be placed into the large bag on the right, with the eggs stacked on the apples.
    Which paper bag contains which object cannot be identified from the initial state. The robot has to search for and scan the barcode on one surface of the bag. }
    \vspace{-0.1in}
    \label{fig:demo}
\end{wrapfigure}
In this demo, we show that our framework can deal with yet another kind of partial grounding in task and motion planning problems.
Initially, we have two paper bags containing different kind of groceries (see Fig.~\ref{fig:demo}). 
The goal is to move them into the large paper bag. To avoid crushing the eggs, the bag of apples cannot be dropped on top of the eggs.
Since the two bags look the same, the robot cannot distinguish between them. We devise a barcode which can be scanned to deduce whether it is eggs or apples.  
We used the following behavior, implemented as a BT, to enable the robot to check and grasp a bag.

\begin{mybox}

\textbf{\pickpointcloud\ Behavior\vspace{-0.08in}}

\tcblower

\indent\textbf{1.} Segment out the point cloud of one paper bag.

\indent\textbf{2.} Compute grasp from point cloud using GPD~\cite{atenpas2017gpd}.

\indent\textbf{3.} Move the bag closer to the camera, and keep rotating it until the barcode (AprilTag~\cite{olson2011apriltag}) is detected.

\indent\textbf{4.} If the barcode matches the target object, the grasp action succeeds, otherwise put this bag back and repeat the above for the next point cloud cluster.

\end{mybox}
This behavior
serves as a local policy that 
fully grounds \pick\text{} during execution. The proposed framework successfully completes the task by using the behavior.

\begin{table*}[ht]
\caption{Horizontal Stacking Benchmark (BE is before execution commences, AE is after execution commences)}
\vspace{-0.1in}
\centering
\setlength{\tabcolsep}{1pt}
\renewcommand{\arraystretch}{1.5}
\scriptsize
\newcolumntype{g}{>{\columncolor{lightgray}}c}
\begin{tabular}{r|g|c|g|c|g|c|g|c|g|c|g|c|g|c|g|c|g|c|g|c|g|c|}
\multicolumn{1}{c|}{}      & \multicolumn{2}{c|}{\textbf{Problem 1}}            & \multicolumn{2}{c|}{\textbf{Problem 2}}            & \multicolumn{2}{c|}{\textbf{Problem 3}}            & \multicolumn{2}{c|}{\textbf{Problem 4}}            & \multicolumn{2}{c|}{\textbf{Problem 5}}            & \multicolumn{2}{c|}{\textbf{Problem 6}}            & \multicolumn{2}{c|}{\textbf{Problem 7}}            & \multicolumn{2}{c|}{\textbf{Problem 8}}            & \multicolumn{2}{c|}{\textbf{Problem 9}}            & \multicolumn{2}{c|}{\textbf{Problem 10}}           & \multicolumn{2}{c|}{\textbf{Mean}}                 \\ \cline{2-23} 
\multicolumn{1}{c|}{}  & {\textbf{Ours}} & \textbf{Base} & {\textbf{Ours}} & \textbf{Base} & {\textbf{Ours}} & \textbf{Base} & {\textbf{Ours}} & \textbf{Base} & {\textbf{Ours}} & \textbf{Base} & {\textbf{Ours}} & \textbf{Base} & {\textbf{Ours}} & \textbf{Base} & {\textbf{Ours}} & \textbf{Base} & {\textbf{Ours}} & \textbf{Base} & {\textbf{Ours}} & \textbf{Base} & {\textbf{Ours}} & \textbf{Base} \\ \hline
{\textbf{Plan-BE (s)}}  & {0.83}          & 0.14          & 0.69          & 0.14          & 7.50          & 3.24          & 9.42          & 0.24          & 0.76          & 0.12          & 0.74          & 0.09          & 6.91          & 3.27          & 7.28          & 3.21          & 7.42          & 0.23          & 0.74          & 0.12          & 4.23          & 1.08          \\ \hline
{\textbf{Plan-AE (s)}}   & 9.22          & 21.83         & 9.15          & 17.62         & 5.01          & 18.61         & 0.18          & 2.30          & 9.07          & 17.18         & 8.93          & 23.15         & 5.06          & 18.08         & 0.12          & 19.59         & 0.16          & 2.34          & 6.91          & 7.86          & 5.38          & 14.86         \\ \hline
{\textbf{Sense-AE (s)}} & 5.38          & 7.06          & 6.41          & 3.54          & 2.36          & 3.42          & 0.82          & 3.75          & 6.41          & 6.68          & 6.17          & 3.64          & 2.29          & 2.71          & 0.77          & 3.18          & 1.02          & 3.04          & 4.88          & 4.20          & 3.65          & 4.12          \\ \hline
{\textbf{Exec (s)}}                   & 94.23         & 172.68        & 95.19         & 110.94        & 94.37         & 144.19        & 75.19         & 99.34         & 106.65        & 166.67        & 95.39         & 112.74        & 98.32         & 151.05        & 82.46         & 134.49        & 76.77         & 92.87         & 81.37         & 94.65         & 89.99         & 127.96        \\ \hline

{\textbf{Total (s)}} & 109.66 & 201.71 & 111.44 & 132.24 & 109.24 & 169.46 &  85.61 & 105.63 & 122.89 & 190.65 & 111.23 & 139.62 & 112.58 & 175.11 &  90.63 & 160.47 &  85.37 &  98.48 &  93.9 &  106.83 & 103.25 & 148.02 \\ \hline

{\textbf{\#Grasps}}              & 5.00          & 9.00          & 5.00          & 6.00          & 5.00          & 8.00          & 4.00          & 5.00          & 5.00          & 9.00          & 5.00          & 6.00          & 5.00          & 8.00          & 4.00          & 8.00          & 4.00          & 5.00          & 4.00          & 5.00          & 4.6           & 6.9           \\ \hline
{\textbf{Success}}              & \cmark          & \cmark          & \cmark          & \cmark          & \cmark          & \cmark          & \cmark          & \cmark          & \cmark          & \cmark          & \cmark          & \cmark          & \cmark          & \cmark          & \cmark          & \cmark          & \cmark          & \cmark          & \cmark          & \cmark          & \cmark           & \cmark           \\ \hline
\end{tabular}
\label{fig:table1}
\end{table*}

\begin{table*}[ht]
\caption{Kitchen Arrangement Benchmark (BE is before execution commences, AE is after execution commences)}
\vspace{-0.1in}
\centering
\setlength{\tabcolsep}{1.5pt}
\renewcommand{\arraystretch}{1.5}
\scriptsize
\newcolumntype{g}{>{\columncolor{lightgray}}c}
\begin{tabular}{r|g|c|g|c|g|c|g|c|g|c|g|c|g|c|g|c|g|c|g|c|g|c|}
\multicolumn{1}{c|}{}      & \multicolumn{2}{c|}{\textbf{Problem 1}}            & \multicolumn{2}{c|}{\textbf{Problem 2}}            & \multicolumn{2}{c|}{\textbf{Problem 3}}            & \multicolumn{2}{c|}{\textbf{Problem 4}}            & \multicolumn{2}{c|}{\textbf{Problem 5}}            & \multicolumn{2}{c|}{\textbf{Problem 6}}            & \multicolumn{2}{c|}{\textbf{Problem 7}}            & \multicolumn{2}{c|}{\textbf{Problem 8}}            & \multicolumn{2}{c|}{\textbf{Problem 9}}            & \multicolumn{2}{c|}{\textbf{Problem 10}}           & \multicolumn{2}{c|}{\textbf{Mean}}                 \\ \cline{2-23} 
\multicolumn{1}{c|}{}  & {\textbf{Ours}} & \textbf{Base} & {\textbf{Ours}} & \textbf{Base} & {\textbf{Ours}} & \textbf{Base} & {\textbf{Ours}} & \textbf{Base} & {\textbf{Ours}} & \textbf{Base} & {\textbf{Ours}} & \textbf{Base} & {\textbf{Ours}} & \textbf{Base} & {\textbf{Ours}} & \textbf{Base} & {\textbf{Ours}} & \textbf{Base} & {\textbf{Ours}} & \textbf{Base} & {\textbf{Ours}} & \textbf{Base} \\ \hline
{\textbf{Plan-BE (s)}} & 7.97 & 11.60 & 6.37 & 10.37 & 8.08 & 12.80 & 9.59 & 11.67 & 11.11 & 9.00 & 41.37 & 14.63 & 35.91 & 12.49 & 34.18 & 23.01 & 12.56 & 7.61 & 12.52 & 8.08 & 17.97 & 12.13 \\ \hline
{\textbf{Plan-AE (s)}} & 26.71 & 0.00 & 23.46 & 0.08 & 32.41 & 0.08 & 0.30 & 0.00 & 0.27 & 0.00 & 1.40 & 0.08 & 0.35 & 0.10 & 0.32 & 0.08 & 0.29 & 0.00 & 1.63 & 0.00 & 8.71 & 0.04 \\ \hline
{\textbf{Sense-AE (s)}} & 6.56 & 0.00 & 5.28 & 0.00 & 5.64 & 0.00 & 2.66 & 0.00 & 1.98 & 0.00 & 2.55 & 0.00 & 2.61 & 0.00 & 2.37 & 0.00 & 2.83 & 0.00 & 2.41 & 0.00 & 3.49 & 0.00 \\ \hline
{\textbf{Exec (s)}} & 91.09 & 23.68 & 96.61 & 30.92 & 95.92 & 29.46 & 102.41 & 13.63 & 106.99 & 22.30 & 96.51 & 59.05 & 95.48 & 57.81 & 95.19 & 69.87 & 114.84 & 13.58 & 108.40 & 13.79 & 100.34 & 33.41 \\ \hline
{\textbf{Total (s)}} &  132.33 &  35.28 & 131.72 &  41.37 & 142.05 &  42.34 & 114.96 &  25.3 &  120.35 &  31.3 & 141.83 &  73.76 & 134.35 &  70.4 &  132.06 &  92.96 & 130.52 &  21.19 & 124.96 &  21.87 & 130.51 &  45.58    \\ \hline
{\textbf{\#Grasps}} & 5.00 & 1.00 & 5.00 & 1.00 & 5.00 & 1.00 & 5.00 & 1.00 & 5.00 & 1.00 & 4.00 & 2.00 & 4.00 & 2.00 & 4.00 & 2.00 & 5.00 & 1.00 & 5.00 & 1.00 & 4.7 & 1.3 \\ \hline
{\textbf{Success}}              & \cmark          & \xmark          & \cmark          & \xmark          & \cmark          & \xmark          & \cmark          & \xmark          & \cmark          & \xmark          & \cmark          & \xmark          & \cmark          & \xmark          & \cmark          & \xmark          & \cmark          & \xmark          & \cmark          & \xmark          & \cmark           & \xmark           \\ \hline
\end{tabular}

\label{fig:table2}
\end{table*}

\subsection{Real-Robot \tamp and Execution Dataset}
We provide our real world benchmarking pipeline as well as a dataset of our runs of all the benchmarking problems~\footnote{Available at: \url{https://github.com/KavrakiLab/tamper-data}}.
The Robot Operating System (ROS)~\cite{Quigley09ros} is required to run our pipeline.
We built a pipeline to record the poses of all the involved objects and obstacles that are relative to the robot frame. To replicate a problem with the real robot, we can visualize 
the initial poses of the objects in RViz~\cite{kam2015rviz} in the robot frame. For any object without an accurate model, a bounding box is necessary to replicate its pose. Finally, we need to manually move around the real-world objects and/or the robot to match the published objects. 
We recorded the real time running data of our method on the benchmarking problems discussed in this section. The data are ROS-bags that can be replayed in simulation.

\section{Discussion}
\label{sec:discussion}

\begin{figure*}[ht]
    \centering
    \includegraphics[trim={0.14in 0 0 0},clip,width=\textwidth]{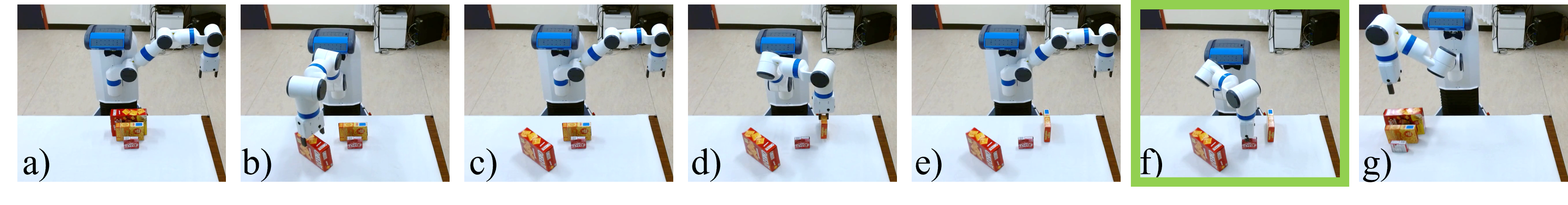}

    \vspace{-0.1in}
    \caption{An example run of the proposed method on the Horizontal Stacking benchmark. A successful use of an online behavior is highlighted in green. \textbf{a)} The robot plans with only sensing the front object, discovers the geometric constraints of the goal \revision{(i.e., planning with the object models, the task and motion planner finds that a smaller object cannot be placed to its goal after a bigger object is already placed there)},
    and finds out that the behavior of picking up the other two objects fails from this state \revision{because they cannot be observed by the head camera}.
    \textbf{b)} The robot replans with the constraints and executes the first action of moving the front object. 
    \textbf{c)} The behavior of picking up the small object fails again in execution \revision{because it is still occluded by the middle object and cannot be observed by the head camera}.
    \textbf{d)} The robot replans again, and moves the middle object away.
    \revision{\textbf{e)} The start state of the behavior for picking up the small object. Here the small object can be observed by the head camera.}
    \textbf{f)} The online execution of the behavior for picking up the small object \revision{succeeds and the behavior finishes with the robot grasping the object.}
    \textbf{g)} \revision{The robot successfully rejoins to the pre-computed trajectory of the next \place\text{} action. Then it proceeds to} the open-loop execution of pre-computed trajectories to move the other objects to the goal.
    }
    \vspace{-0.05in}
    \label{fig:h3_pane}
\end{figure*}

\begin{figure*}[ht]
    \centering
    \includegraphics[trim={0.14in 0 0 0},clip,width=\textwidth]{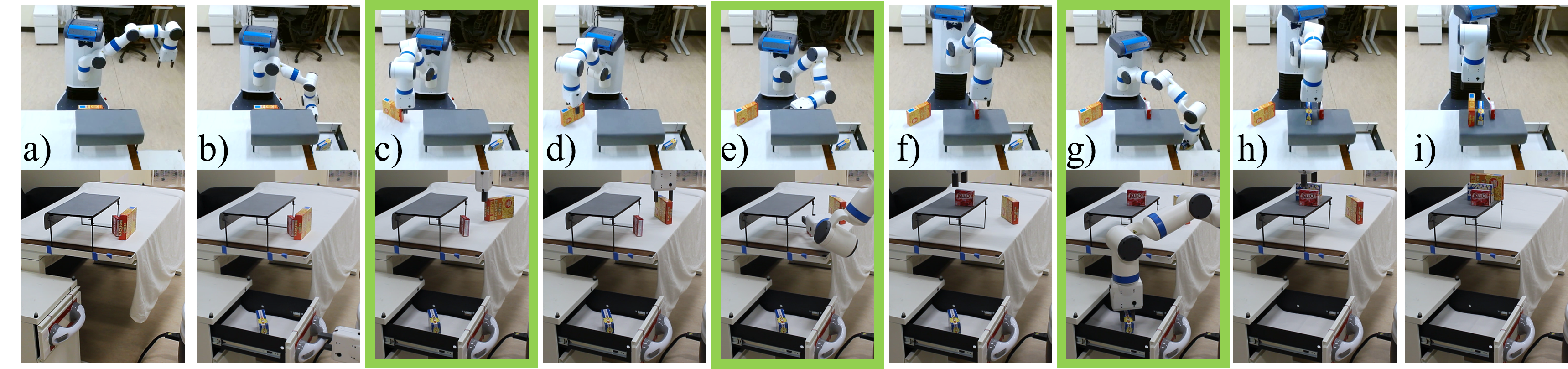}
 
    \vspace{-0.1in}
    \caption{An example run of the proposed method on the Kitchen Arrangement benchmark. The successful uses of an online behavior are highlighted in green. 
    \textbf{a)} The robot plans with only sensing the front object,  discovers the geometric constraints of the goal, and finds out that the behavior of picking up the small object fails from this state. 
    \textbf{b)} The robot opens the drawer. 
    \textbf{c)} The behavior of picking up the small object fails again in execution, as it is still not revealed. The robot then replans and executes the online behavior of pushing and picking up the front object.
    \textbf{d)} The robot executes the pre-computed trajectory of placing it at an intermediate location.
    \textbf{e)} Online execution of the behavior for picking up the small object which is now revealed.
    \textbf{f)} Open-loop execution of placing the small object on the shelf.
    \textbf{g)} Online execution of the behavior of picking up the object from the drawer.
    \textbf{h-i)} Open-loop execution of pre-computed trajectories to move the objects to the goal.
    }
    \vspace{-0.1in}
    \label{fig:ds_pane}
\end{figure*}

This work introduces a task and motion planning for execution in the real (\acronym) framework that addresses realistic scenarios where the symbolic domain might be known but the geometric knowledge of the world has gaps which often cannot be resolved before execution begins.
This work has pushed the notion of the \tamp paradigm into the execution layer by addressing the incomplete \tamp domain during planning. The power of traditional task planning and motion planning is applied wherever information is known, and a partially grounded task and motion plan is constructed with gaps.
These gaps are bridged during execution using closed-loop behaviors, which are assumed to be input and can be implemented using hand-crafted subroutines, behavior trees, or learned controllers. Information can then be fed back as execution constraints to the framework to potentially replan.

The \acronym framework overcomes the limitations of traditional \tamp solvers, which need all the information about the \tamp domain to be known during planning, as well as more recent techniques which either decouple the task-motion problem, or only address fully defined but partial \tamp domains incrementally using replanning.
The real-world data is accumulated during our experiments is also being shared in the form of a dataset to allow broader accessibility to realistic problems and operational sensing data, and enable researchers to recreate or evaluate the studied problems.

The current work also opens up a lot of interesting points of discussion. The proposed framework has been presented from the point of view of expanding the domain of applicability of traditional \tamp methods, which have grown popular as a powerful and general long-horizon planning paradigm. To endow the power of execution robustness, the availability of specialized behaviors is assumed as input. Although this has been demonstrated to be reasonable for a suite of benchmarks and demonstrations, it is definitely an open problem---
how can behaviors be generated automatically;
\revision{how can we actively discover the needs of generating the behaviors during real-world execution;}
how can behaviors be evaluated; what is the best way to model behaviors; how many behaviors are sufficient? The current work opens the door to addressing these questions in the future towards more complex compositional, extensible, and targeted robotic frameworks that are designed to be effective in the real world.

Our framework shows a trade-off between re-computing a new task plan upon grounding failure and leaving it as a gap in the task and motion plan to be filled by a behavior. We choose to prioritize using a behavior to fill the grounding gap rather than finding a new task plan. 
Note that although the execution of the behaviors is not guaranteed to succeed, such constraints can only be discovered after real-world executions, and discovering them may be crucial to eventually solving the problem.
However, there can be cases where computing and grounding an alternate task plan is more efficient than always committing to the behavior.
For example, if grasping any of the two objects fulfills the goal, where one is occluded and one is not, grounding the alternate task plan of grasping the visible object is much more efficient than trying to use a behavior to grasp the occluded object and failing.
Whether to prioritize planning or execution effort depends on the specific problem setting. If the behaviors are expected to be reliable or that specific action is critical to the feasibility, it might be better to always try executing a behavior. 
If there are several feasible task plans, alternative solutions that do not involve behaviors can be searched. However, this can be expensive in itself and is liable to lead to suboptimal task plans. This trade-off is subject to the needs of the problem.

Another question is the modeling of uncertainty. Gaps can be seen from the lens of instances of belief uncertainty modeled in the world.
It is of definite interest to explore 
how more informed modeling of the underlying uncertainty can be better utilized while preserving performance. 
For example, even though it still remains an open problem how to efficiently model long-horizon manipulation tasks as POMDPs, it would be interesting to explore how can we combine the proposed framework which focuses on long-horizon tasks, with the general POMDP formulation to deal with uncertainty in the real world in an informed and structured way.

The present work serves as an augmentation of traditional algorithmic planning approaches into realistic domains, which have been recently dominated by learning-based techniques, due to their robustness to real sensing data. Through the behaviors used in the current work, hooks for incorporating powerful learning-based grounding mechanisms as well as local controllers are provided within a high-level framework. This opens up investigations into the role of learning-based methods when designed specifically for the proposed paradigm of robots that reason about tasks, motions, and execution.

An important aspect of traditional \tamp methods has been their theoretical soundness and guarantees. Typical asymptotic guarantees which practically demonstrate convergence properties to either feasibility or optimality become ill-posed when execution in real-time is introduced.
\revision{As an example, typical \tamp methods can achieve probabilistic completeness, avoiding unrecoverable states by e.g., backtracking~\cite{dantam2018incremental, garrett2020pddlstream, shah2020anytime, srivastava2014combined}. Such states have to be modeled in the problem domain to allow reasoning over them. If execution failures happen leading to unmodeled dead-end states (e.g., glass cup fallen and broken to pieces), the typical TAMP methods cannot recover and have to report failure. In this sense, we hold similar assumptions that our framework avoids the modeled catastrophic failures by leveraging the provided behaviors (e.g., the Kitchen Arrangement benchmark in Fig.~\ref{fig:catastrophic}). If dead-end states exist that are not modeled by our \tamp module or by the behaviors, we can also only report failure, similarly to typical \tamp methods.
On the other hand, if we assume that the execution of all behaviors do not lead to irreversible states in the real-world, the proposed framework achieves probabilistic completeness if all the planning modules used are probabilistic complete. 
}
For the planning and execution problem as a whole to express reasonable guarantees, there is a need to rethink the kind of finite-time properties we can leverage out of traditional techniques.

The wealth of exciting directions the present work opens up is testament to the promise held by pushing the applicability of robotic long-horizon reasoning to completing tasks within real world execution. The proposed \acronym framework is a significant stepping stone towards achieving this goal.

\bibliography{references-cleaned.bib}

\vfill

\end{document}